\documentclass[10pt,journal,compsoc,utf8]{IEEEtran}
\usepackage{makecell,interfaces-makecell}
\usepackage{tabu}
\usepackage{graphicx}
\usepackage{amssymb}
\usepackage{booktabs}
\usepackage{tabularx}
\usepackage{multirow}
\usepackage{xcolor}
\usepackage{printlen}
\usepackage{graphbox}
\usepackage{colortbl}
\usepackage{microtype}
\usepackage{mathtools}
\usepackage{url}
\usepackage[absolute,overlay]{textpos}
\usepackage{xspace}

\newif\ifarxiv
\arxivtrue

\ifarxiv
\usepackage{hyperref}
\fi

\ifCLASSOPTIONcompsoc
  \usepackage[nocompress]{cite}
\else
  \usepackage{cite}
\fi

\ifCLASSINFOpdf
  \DeclareGraphicsExtensions{.pdf,.jpeg,.jpg,.png}
\else
\fi

\usepackage{amsmath}
\makeatletter
\DeclareRobustCommand\onedot{\futurelet\@let@token\@onedot}
\def\@onedot{\ifx\@let@token.\else.\null\fi\xspace}
\def\eg{\emph{e.g}\onedot} 
\def\ie{\emph{i.e}\onedot} 
\def\cf{\emph{c.f}\onedot} 
 
\def\wrt{w.r.t\onedot} 
\def\etal{\emph{et al}\onedot}

\definecolor{darkgreen}{HTML}{1b6319}
\def\st{$^{*}$}
\makeatother

\newcommand{\centercell}[1]{\multicolumn{1}{c}{#1}}
\newcommand{\std}[1]{\raisebox{.1\height}{\scalebox{.8}{$\pm$#1}}}
\newcommand{\smalletal}{\scalebox{.8}{\etal}}
\newcommand{\un}[1]{#1}

\begin{document}

\title{MeTRAbs: Metric-Scale Truncation-Robust Heatmaps for Absolute 3D Human Pose Estimation}

\author{Istv\'an~S\'ar\'andi,~\IEEEmembership{Student Member,~IEEE,}
        Timm~Linder,~\IEEEmembership{Member,~IEEE,}\\
        Kai~O.~Arras,~\IEEEmembership{Member,~IEEE,}
        Bastian~Leibe,~\IEEEmembership{Member,~IEEE}%
\IEEEcompsocitemizethanks{
\IEEEcompsocthanksitem I. S\'ar\'andi and B. Leibe are with RWTH Aachen University, Germany. Email: \{Sarandi, Leibe\}@vision.rwth-aachen.de\protect\\
\vspace{-3mm}
\IEEEcompsocthanksitem T. Linder and K. O. Arras are with Robert Bosch GmbH, Renningen, Germany. Email: \{Timm.Linder, KaiOliver.Arras\}@de.bosch.com}%
\thanks{\copyright~2020 IEEE. Personal use of this material is permitted.  Permission from IEEE must be obtained for all other uses, in any current or future media, including reprinting/republishing this material for advertising or promotional purposes, creating new collective works, for resale or redistribution to servers or lists, or reuse of any copyrighted component of this work in other works.}
\thanks{\normalfont{\textsuperscript{1} \texttt{\url{https://vision.rwth-aachen.de/metrabs}}}}
}

\markboth{}%
{S\'ar\'andi \MakeLowercase{\textit{et al.}}: MeTRAbs: Metric-Scale Truncation-Robust Heatmaps for Absolute 3D Human Pose Estimation}

\IEEEtitleabstractindextext{%
\begin{abstract}
Heatmap representations have formed the basis of human pose estimation systems for many years, and their extension to 3D has been a fruitful line of recent research.
This includes 2.5D volumetric heatmaps, whose X and Y axes correspond to image space and Z to metric depth around the subject.
To obtain metric-scale predictions, 2.5D methods need a separate post-processing step to resolve scale ambiguity.
Further, they cannot localize body joints outside the image boundaries, leading to incomplete estimates for truncated images.
To address these limitations, we propose metric-scale truncation-robust (\emph{MeTRo}) volumetric heatmaps, whose dimensions are all defined in metric 3D space, instead of being aligned with image space.
This reinterpretation of heatmap dimensions allows us to directly estimate complete, metric-scale poses without test-time knowledge of distance or relying on anthropometric heuristics, such as bone lengths.
To further demonstrate the utility our representation, we present a differentiable combination of our 3D metric-scale heatmaps with 2D image-space ones to estimate absolute 3D pose (our \emph{MeTRAbs} architecture).
We find that supervision via absolute pose loss is crucial for accurate non-root-relative localization.
Using a ResNet-50 backbone without further learned layers, we obtain state-of-the-art results on Human3.6M, MPI-INF-3DHP and MuPoTS-3D.
Our code is publicly available.\textsuperscript{1}
\end{abstract}

}

\maketitle

\IEEEdisplaynontitleabstractindextext

\IEEEpeerreviewmaketitle

\IEEEraisesectionheading{%
\section{Introduction}\label{sec:introduction}}
\IEEEPARstart{H}{uman} pose estimation is a long-standing computer vision problem with wide applicability in human-robot interaction~\cite{Zimmermann18ICRA}, virtual reality~\cite{Alldieck2018CVPR}, medicine~\cite{Srivastav18MICCAIW} and commerce~\cite{Neverova18ECCV}, among others.
Since the adoption of deep convolutional neural networks (CNN), and especially heatmap representations, we have witnessed rapid progress in pose estimation research~\cite{Newell16ECCV,Yang17CVPR,Ke18ECCV}.
A particularly challenging task is monocular 3D pose estimation~\cite{Martinez17ICCV,Mehta17TOG,Zhou17ICCV,Luo18BMVC,Nibali19WACV}, where a person's anatomical landmarks are sought in 3D space, \ie, in millimeters, instead of pixels, given only a single image.
Reconstructing 3D from images is one of the major goals of computer vision research, but several geometric ambiguities make this challenging.
First, different 3D articulations may share the same 2D projection and second, there is an ambiguity between object size and distance, as small objects near the camera appear the same as large ones far away.

\begin{figure}[!t]
\centering
\includegraphics[height=51mm]{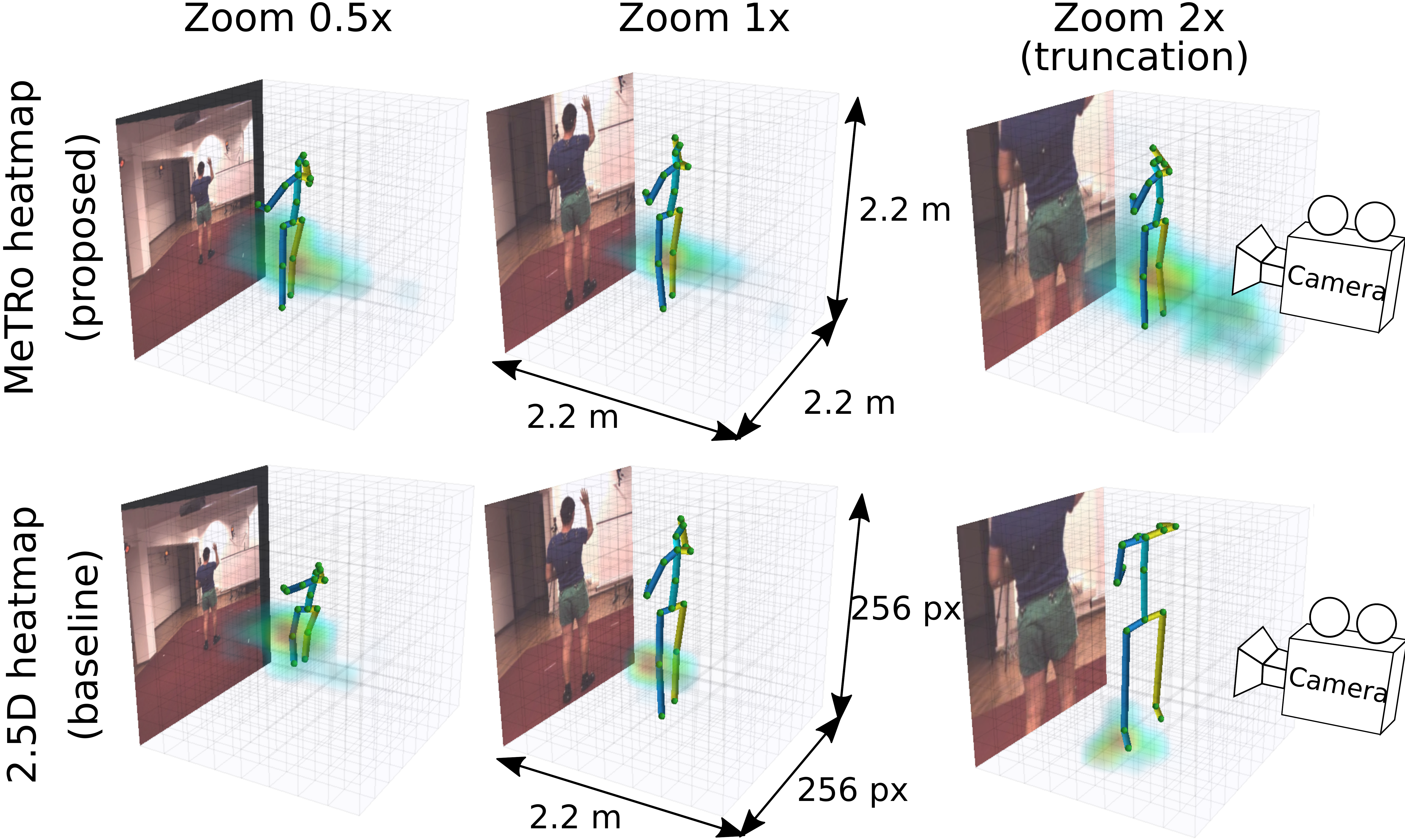}\\
\caption{By defining heatmaps in the 3D metric space around the person (\emph{top row}) we directly estimate scale-correct complete poses.
This is in contrast to prior work (\emph{bottom row}) that defines the $X$ and $Y$ heatmap axes in image space and requires further post-processing to obtain a metric-scale skeleton.
The three columns show that this heatmap representation is nearly invariant \wrt image zooming.
A knee heatmap is shown along with the soft-argmax decoded skeleton.
}
\label{fig:volume_visu}
\end{figure}
There is no clear consensus on the most effective way to represent and tackle these problems.
Heatmap estimation is a promising direction, because it makes direct use of the convolutional structure of CNNs by turning the coordinate estimation problem into a binary classification problem of whether the joint is located at the given position or not.
To estimate 3D pose, a successful line of works extends 2D joint heatmaps with a depth axis, resulting in a 2.5D volumetric representation~\cite{Pavlakos17CVPR,Sun18ECCV,Iqbal18ECCV,Luvizon18CVPR}.

Finding heatmap maxima gives the estimated pixel coordinates and root-relative metric depths per joint (a 2.5D pose).
While these estimates can be accurate, 2.5D representations do not address the challenging ambiguity between person size and distance.
To bridge the gap between a 2.5D and a 3D pose, a separate scale recovery step is needed in post-processing.
Explicit anthropometric heuristics have been proposed as scale cues, \eg bone length priors~\cite{Pavlakos17CVPR} or a skeleton length prior~\cite{Sun18Arxiv}, computed by averaging over the training poses.
However, these simple heuristics have difficulties when the experimental subjects have diverse heights.
A further limitation is that 2.5D formulations are constrained to the estimation of joints that lie within the image boundaries.
This is problematic in practical applications, where the image crop may not include the whole person, \eg due to occlusions or detector noise.
While one could use an additional module to fill in missing joints, it is desirable to learn the complete skeleton estimation in a single unified stage.

Our goal in this paper is to tackle scale and distance estimation of 3D poses in a truncation-robust, simple and efficient manner, while also keeping the structural advantages of fully-convolutional heatmap estimation, as opposed to numerical coordinate regression (\ie encoding position by activation \emph{location} instead of activation \emph{value}).

To this end, we propose training a fully-convolutional network to output our novel metric-scale truncation-robust (MeTRo) heatmaps as illustrated in Fig.~\ref{fig:volume_visu}.
All dimensions of these heatmaps are defined to have a fixed metric extent in meters.
This is an unconventional task definition for fully-convolutional networks (FCN).
FCNs are predominantly applied for pixel-wise prediction tasks, such as semantic segmentation, where the input and output are pixel-to-pixel aligned.
In our proposed approach, the input pixel positions and the output metric positions only satisfy a looser form of spatial correspondence.
Nevertheless, we show that somewhat surprisingly, such a mapping can still be learned effectively by a standard modern FCN backbone.

By skipping the 2.5D stage, the backbone FCN has to implicitly reason about out-of-image joints, discover scale cues and learn the geometric perspective back-projection in an end-to-end manner.
Our MeTRo heatmaps can naturally encode body parts outside the image, since the prediction volume's bounds do not correspond to the image bounds.
As there is no need to design an explicit scale recovery step, the pipeline becomes simpler and requires neither the focal length nor the root joint distance to be known at test time for root-relative prediction.

\begin{figure*}[!t]
\centering
\includegraphics[width=\textwidth]{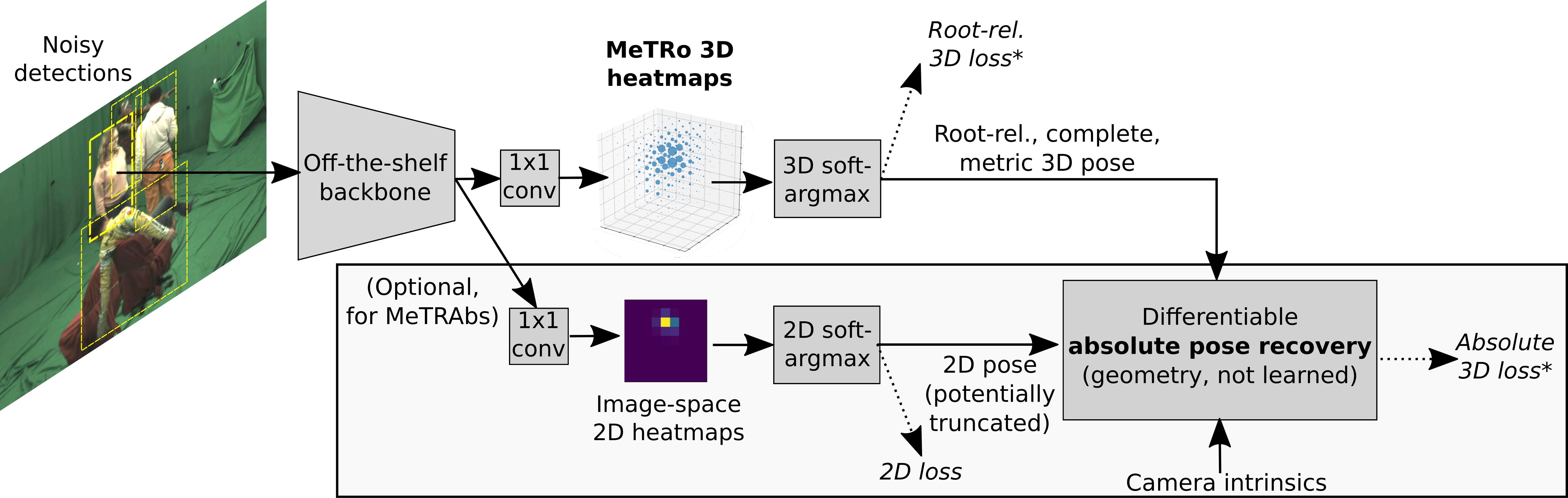} \\
\caption{Overview of our approach.
We predict volumetric heatmaps using an off-the-shelf fully-convolutional backbone.
Applying soft-argmax on these heatmaps and scaling by an image-independent constant factor yields joint coordinates in metric space up to translation.
We minimize the root-relative $L^1$ loss.
Focusing on simplicity, no learnable parameters are introduced outside the standard backbone, except for a single 1$\times$1 convolution.
Optionally, if absolute (non-root-relative) pose estimation is required, our \textit{MeTRAbs} extension also estimates classic 2D image-space heatmaps via another 1$\times$1 convolutional head.
We then reconstruct the absolute pose through a differentiable reconstruction module.
This is based on a linear least squares formulation derived from the pinhole camera model.
Supervision is applied both at the outputs of the individual prediction heads and at the final combined output.
(\st For 2D-labeled examples, the root-relative loss is replaced by a scale and translation-invariant 2D loss and the absolute 3D loss is not used.)}
\label{fig:absolute_pose}
\end{figure*}

Employing the differentiable soft-argmax~\cite{Levine16JMLR,Luvizon18CVPR,Sun18ECCV,Nibali18arXiv} layer, our method becomes end-to-end learned all the way from image to final 3D metric-scale prediction as shown in Fig.~\ref{fig:absolute_pose}.
Soft-argmax also allows rapid training with low-resolution heatmaps.
Without any additional learned decoder module, we perform dense prediction with reduced strides at test time for higher quality results.
We find that the details of the striding mechanism are crucial and propose a \textit{centered striding} method that distributes the output neuron receptive fields evenly over the image.

This paper presents an extension of our own previous work~\cite{Sarandi20FG}.
While in~\cite{Sarandi20FG} we only considered single-person root-relative pose, here we show that MeTRo heatmaps are also effective for absolute (non-root-relative) 3D pose estimation.
In multi-person scenes it is especially important to estimate absolute poses, in order to recover the spatial layout of the whole group.
We combine 3D metric-scale root-relative heatmaps with 2D image-space heatmaps in a two-headed CNN architecture, and subsequently reconstruct the absolute 3D root position in a differentiable manner.
While prior approaches have tackled the root reconstruction problem, to our knowledge we are the first to backpropagate gradients through this reconstruction, allowing us to end-to-end supervise the absolute pose task.
We evaluate our network in a top-down fashion combined with an off-the-shelf person detector.
We refer to this combined approach as MeTRAbs.

Recent approaches have achieved good generalization performance to in-the-wild images by using abundant and diverse images with 2D pose labels in the training procedure besides 3D data~\cite{Zhou17ICCV,Sun18ECCV,Luvizon18CVPR}.
Applying such weak supervision is challenging in our representation, since the MeTRo heatmap would require supervision with metric ground truth instead of the image-space ground truth supplied with 2D datasets.
We tackle this by proposing a scale and translation agnostic loss for 2D-annotated examples using an alignment layer.
Note that in contrast to 2.5D heatmaps, this alignment is only used for loss computation during training, and still allows MeTRo to infer joints outside the image boundaries.

Experimentally, we achieve state-of-the-art results on the two largest single-person 3D pose benchmarks, Human3.6M and MPI-INF-3DHP, as well as the popular multi-person dataset MuPoTS-3D.
To isolate the effect of the representation, we perform direct comparisons with 2.5D heatmap learning using bone-length-based scale recovery~\cite{Pavlakos17CVPR}, under otherwise equal training conditions.
We find that scale cues can indeed be learned implicitly in this fashion and MeTRo outperforms the baseline on most test sequences.
Furthermore, our approach achieved first place in the 2020 ECCV 3D Poses in the Wild~\cite{VonMarcard18ECCV} Challenge.

In summary, we make the following contributions: 
1) We propose a novel 3D heatmap representation for pose estimation, called \textit{MeTRo}, whose dimensions are defined in a fixed metric scale, irrespective of the input image scale.
We achieve state-of-the-art results on Human3.6M and MPI-INF-3DHP and demonstrate strong truncation-robustness.
2) We propose \emph{centered striding}, an improvement to the usual CNN striding logic, enabling higher accuracy at a coarse (8$\times$8) heatmap resolution.
3) For absolute pose estimation, we extend the MeTRo approach to \textit{MeTRAbs}, by also estimating 2D image-space heatmaps from the same backbone and reconstructing the absolute pose.
We achieve state-of-the-art results on the MuPoTS-3D and 3DPW multi-person benchmarks using MeTRAbs in a top-down paradigm.
4) To our knowledge, we are first to use a monocular geometry-based differentiable absolute pose reconstruction module to supervise the network with the final absolute ground truth fully end-to-end.
We show that this is crucial for good distance estimation and extensively evaluate strong and weak perspective-based reconstruction variants.
\section{Related Work}
\label{sec:related}
3D human pose estimation has had a long research history starting with hand-crafted features and part-based models~\cite{Sarafianos16CVIU}.
Similar to other computer vision problems, the transition to deep convolutional networks has led to a dramatic performance increase in this task.
For a thorough overview, see the recent survey by Chen \etal~\cite{Chen20CVIU}.

\subsection{Deep 3D Human Pose Estimation}
Much of the inspiration in recent 3D pose estimator design has come from lessons learned in 2D pose research.
DeepPose, the first neural 2D pose estimator~\cite{Toshev14CVPR} directly regressed 2D joint coordinates on the RGB input via convolutional and fully-connected layers.
Later, top-performing 2D methods have transitioned to predicting body joint heatmaps by fully-convolutional networks (\eg,~\cite{Newell16ECCV}) as an intermediate representation.
These heatmaps are spatially discretized arrays (one for each joint), in which higher values indicate higher confidence that the particular joint is located at the corresponding position.

One line of 3D pose research builds on top of 2D heatmaps and infers the 3D pose from them by exemplar-matching~\cite{Chen17CVPR}, regression~\cite{Martinez17ICCV} or probabilistic inference~\cite{Tome17CVPR}.
One downside of such approaches is that the image content only indirectly influences the 3D estimation, as it acts on the result of the 2D estimation stage.
Furthermore, 2D-to-3D lifting is performed in a numerical coordinate representation, which does not benefit from the built-in convolutional structure of CNNs.

Nibali \etal~\cite{Nibali19WACV} predict three marginal heatmaps per body joint, for the XY, XZ and YZ planes, respectively.
Pavlakos \etal~\cite{Pavlakos17CVPR} propose extending 2D heatmaps with a root-relative metric depth axis.
One can obtain the 2D pixel positions and root-relative depths of the joints by finding maxima in the heatmaps.

One downside of heatmap representations has been the requirement of a dense output, which can become especially costly in 3D.
The recently proposed soft-argmax~\cite{Levine16JMLR,Luvizon19CG,Nibali18arXiv} \emph{a.k.a.} integral regression~\cite{Sun18ECCV} method greatly alleviates this problem.
As opposed to hard-argmax, which simply finds the location of the highest heatmap activation, soft-argmax is computed as the weighted average of all voxel grid coordinates, using softmaxed heatmap activations as the weights.
For example, a low resolution heatmap can encode a joint position lying halfway between two bin centers by outputting 0.5 for both bins.
By virtue of being differentiable, unlike hard-argmax, it also obviates the need for explicit heatmap-level supervision (\eg, voxel-wise cross-entropy).
Instead, the loss can be computed (and its gradients back-propagated) from the coordinates yielded by soft-argmax.

Besides 2D heatmaps, Mehta \etal~\cite{Mehta17TOG} estimate three further output channels per joint, the so-called \emph{location maps}.
These are read out at the position of the corresponding heatmap's peak to obtain the X, Y and Z coordinates on a metric scale.
Note how in their approach the final 3D joint coordinates are generated in the form of activation \emph{values} (of the location maps at the heatmap peaks), as opposed to high-activation \textit{locations}.
We can thus think of it a conceptual hybrid of direct numerical coordinate regression and heatmap estimation.
A downside of this method is that it requires high-resolution location maps and cannot benefit from the soft-argmax approach.

\subsection{Scale and Distance Estimation}
It is well-known that projecting a 3D world onto a 2D image plane results in ambiguity between size and distance (depth).
However, the end goal for 3D scene understanding and 3D human pose estimation in particular is a metric-space output at the true scale.
The ambiguity can only be resolved using semantic scale cues, \ie prior knowledge of the usual size of humans and other objects appearing in the scene.
Unfortunately, not all papers describe how this step is performed.
Some authors report their results assuming an already known ground-truth root joint distance and focal length~\cite{Nibali19WACV,Sun18ECCV,Sun18Github,Chen19BMVC}.
A simple anthropometric approach is used by Pavlakos \etal.~\cite{Pavlakos17CVPR}
Given 2D pixel positions and root relative depth estimates from volumetric heatmaps, they optimize the absolute person distance such that the back-projected skeleton's bone lengths match the average over the training set in a least squares sense, assuming a full perspective model.
We use this scale recovery approach as our main baseline comparison throughout the paper, described in more detail in Sec.~\ref{sec:baseline}.
Sun \etal~\cite{Sun18Arxiv} employ a similar idea, but use the overall skeleton length and a weak perspective model instead.
Methods that are not based on volumetric heatmaps~\cite{Rogez19PAMI,Mehta18TDV} can directly predict the metric-scale numerical coordinates.
Some recent works have shown that direct regression of person height from an image is a challenging task~\cite{Gunel18ICCVW,Dantcheva18ICPR}.

While monocular 3D pose estimation methods are typically only evaluated in a root-relative manner, some works have also explicitly tackled the absolute (non-root-relative) pose estimation task, where every joint position is predicted within the 3D camera coordinate frame.
This is closely related to the above-discussed metric-scale prediction: if both the image-space pose and the metric-scale root-relative pose are known, one can reconstruct the absolute distance (assuming a calibrated camera).
Mehta \etal~\cite{Mehta17TDV} and Dabral \etal~\cite{Dabral19TDV} reconstruct the root offset by assuming a weak perspective model.
Mehta \etal~\cite{Mehta19Arxiv} assume the foot touches the known ground plane in the first frame.
Moon \etal~\cite{Moon19ICCV} predict the metric area of the human bounding box as a numerical value via a separate deep network (RootNet), besides their root-relative 2.5D PoseNet.
In contrast to Moon \etal, we estimate the scaled pose fully-convolutionally and do not require multiple separate backbones.
In our earlier work~\cite{Sarandi18Arxiv}, we estimate the distance directly from the image crop, but that does not generalize well to novel environments.
Dabral \etal\cite{Dabral19TDV} propose to estimate the focal length jointly with the distance, implicitly relying on the perspective distortion of people far from the optical axis.
As the authors note, this cannot work well when the camera is turned directly towards the target person.
V\'eges \etal~\cite{Veges19IJCNN} make use of a monocular depth prediction network pretrained on various indoor and outdoor datasets to help with absolute person distance estimation.
Finally, some recent works also consider the depth relations among people: Jiang \etal~\cite{Jiang20CVPR} optimize the depth ordering by occlusion cues, while Fieraru \etal~\cite{Fieraru20CVPR} explicitly localize contact points between people to help with coherent reconstruction.
In contrast, we perform our estimation for each person independently.

\subsection{Truncated Pose Estimation}
Single-person 3D pose estimation benchmarks, such as Human3.6M~\cite{Ionescu11ICCV,Ionescu14PAMI}, assume that the whole person is visible in the input image.
In practical applications, however, bounding boxes are obtained using imperfect detectors, which can result in body truncation, especially in high-occlusion scenes.
A possible remedy could be extending the detection crops by amodal completion~\cite{Kar15ICCV}, but this would result in a loss of image resolution.
Generally, pose estimation performance under truncation has not been studied extensively in the literature.
Recent work by Park \etal~\cite{Park20PRL} uses cropping data augmentation to improve 2D pose estimation.
Vosoughi \etal create randomly truncated crops from Human3.6M images, and show that current methods perform poorly on truncated person images, even when only considering the present (within-boundary) joints~\cite{Vosoughi18ICIP}.
They tackle the problem using direct numerical coordinate regression, similar to early 2D pose estimation methods~\cite{Toshev14CVPR}.
We show that our approach performs significantly better in the truncated setting.
\section{Single-Person Root-Relative Approach}
\label{sec:metro}
In this section we present our proposed approach for metric-scale root-relative 3D human pose estimation.
The input is an RGB image crop $I\in \mathbb{R}^{w\times h\times 3}$ depicting a person.
The desired output is a 3D skeleton, consisting of $J$ joint coordinates $\left\{(\Delta X_j, \Delta Y_j, \Delta Z_j)^T\right\}_{j=1}^{J}$ in millimeters, up to arbitrary translation (hence the $\Delta$ symbols).

\subsection{Metric-Space Volumetric Heatmap Representation}
\begin{figure}[t]
\centering
\includegraphics[width=87mm]{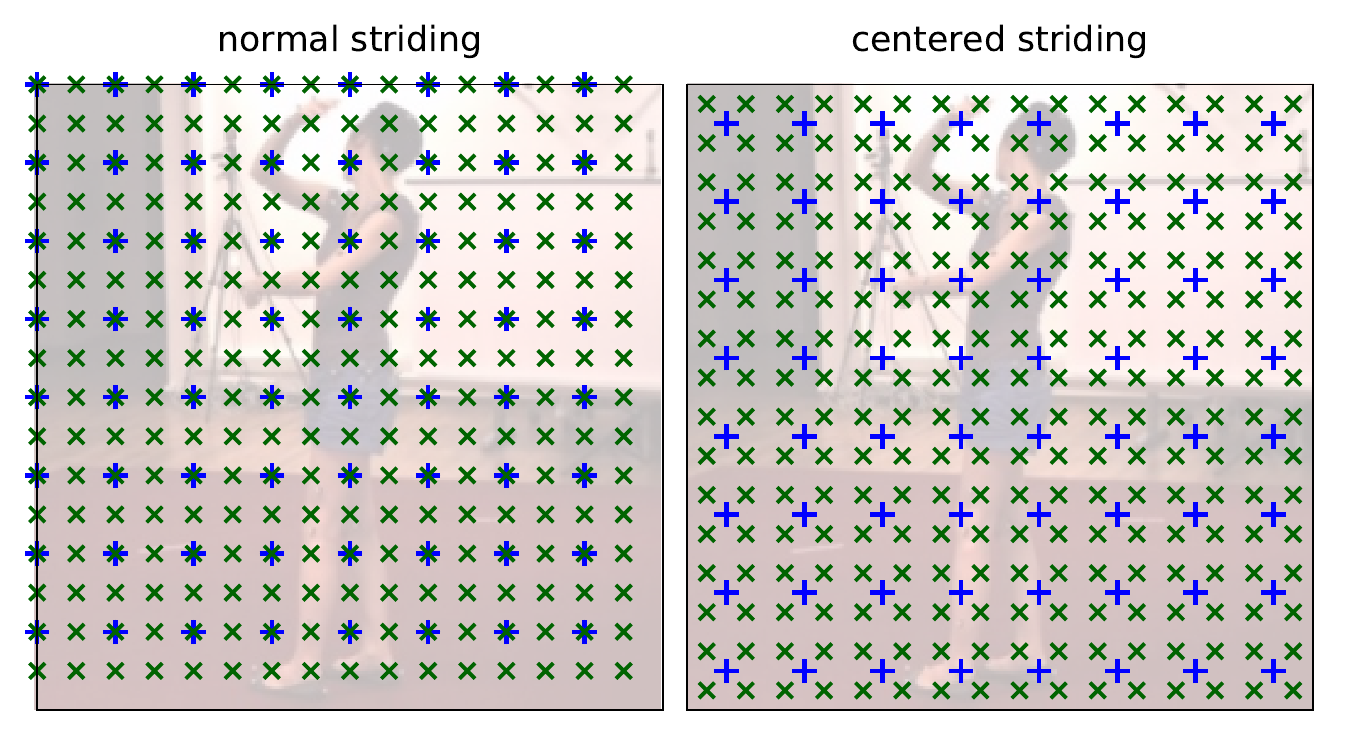}\\
\caption{Receptive field centers of the output neurons in a strided FCN operating on a 256$\times$256 image ({\color{blue} +}: stride 32, {\color{darkgreen}$\times$}: stride 16).
\emph{left}: normal striding logic, where the top left result is kept per 2$\times$2 block.
Note that denser striding skews the sample density towards the bottom and right in the border areas.
\emph{right:} by reversing the stride logic in the last strided layer (\ie, bottom right result taken, instead of top left), the samples are centered and the increased striding density is distributed evenly.
}
\label{fig:striding}
\end{figure}

As is common in heatmap-based approaches, we apply a fully-convolutional backbone network, with effective stride $s$ to produce an array with $d\cdot J$ spatial output channels.
Here $d$ is the number of discretization bins along the depth axis of the prediction volume.
We then split the array along the channel axis into $J$ volumes, each of shape $(w/s) \times (h/s) \times d$.
3D spatial softmax is applied over each of them, resulting in volumetric heatmap activations $V^{(j)} \in \mathbb{R}^{(w/s) \times (h/s) \times d}$.
Up to this point the process is similar to other volumetric heatmap approaches~\cite{Sun18ECCV,Pavlakos17CVPR}.
The difference lies in how the heatmap axes are interpreted to yield metric-scale coordinates.
In particular, the 3D joint coordinates are decoded using soft-argmax with \emph{fixed} scaling factors:
\begin{equation}
\begin{bmatrix}\Delta X_j\\ \Delta Y_j \\ \Delta Z_j \end{bmatrix} = \sum\displaylimits_{p,q,r} V_{p,q,r}^{(j)}\cdot \hspace{-0.5mm} \begin{bmatrix} p \cdot s/w \cdot W \\ q \cdot s/h \cdot H \\ r\cdot 1/d \cdot D \end{bmatrix}, \\
\label{eq:decoding}
\end{equation}
where the $p, q, r$ are 0-based integer indices into the volumetric heatmap array and $W, H, D$ are the fixed metric width, height and depth extents of the full prediction volume.
We set these extents as 2.2 meters in our work, which allows capturing people of usual height even when stretched out.
Depending on striding logic (see Sec.~\ref{sec:stride}), Eq.~\ref{eq:decoding} needs to be adjusted slightly, \eg the volume size may change with denser striding (Fig.~\ref{fig:striding}).
The final root-relative prediction is obtained by subtracting the predicted root coordinates from all joint positions.
Supervision is applied on these root-relative coordinates.
This means that the position of the root joint prediction within the volume is not explicitly supervised and the network can place the skeleton anywhere within the prediction volume.
The gradients are backpropagated through the root-joint-subtraction operation.
No camera calibration-based back-projection, nor bone or skeleton size-based rescaling is needed for this root-relative prediction.
The network is trained to perform these operations implicitly within the backbone.

\subsection{Architecture}
In contrast to prior work that employs decoders with upsampling layers and multiple refinement stages, we show that the task can be tackled in a significantly simpler fashion.
Indeed, we apply the widely used ResNet-50~\cite{He16ECCV} backbone to directly predict spatial heatmaps, without any additional learnable layers, such as transposed convolutions.
By default, ResNet-50 has an effective stride of 32, resulting in heatmaps of spatial size 8$\times$8 from the input image of size 256$\times$256 during training.
The depth of the volumetric heatmap is set to 8.
When testing on single-person datasets, we apply the trained network with an effective stride of 4, to obtain heatmaps with spatial size 64, which is the typical size used in prior work~\cite{Sun18ECCV,Pavlakos17CVPR}.
This is called dense prediction and is commonly used in image segmentation~\cite{Chen18PAMI}.
In this technique, striding is removed from a given number of convolutional layers and the dilation rate of subsequent convolutions is increased correspondingly.
As we will see, dense prediction increases the compute requirements but also improves accuracy, while still allowing real-time execution.
\subsection{Centered Striding}
\label{sec:stride}

When changing striding density at test time compared to training time, it is important to consider how the distribution of heatmap receptive field centers is affected.
The left side of Fig.~\ref{fig:striding} shows a 256$\times$256 image processed with training stride 32 ({\color{blue} +}) and test stride 16 ({\color{darkgreen}$\times$}).
The coverage changes significantly between training and test and is not symmetric over the image.
While not an issue for pixel-labeling tasks, soft-argmax is a weighted vote-averaging scheme and introducing new voting positions in an uneven manner skews the prediction result.
To tackle this issue, we propose \textit{centered striding}, where the striding logic in the last convolutional layer of the backbone is ``reversed'', such that it outputs the \emph{bottom right} result per each 2$\times$2 block.
The result is a more evenly distributed coverage over the image, with each original sampling position replaced with four new ones equally spaced around it.
This benefit is evaluated in Sec.~\ref{sec:results}.
\subsection{Scale and Translation Agnostic 2D Loss}
Similar to recent approaches~\cite{Zhou17ICCV,Sun18ECCV,Luvizon18CVPR}, we train simultaneously on 3D-labeled data from motion capture studios and 2D-labeled, in-the-wild data from the MPII dataset~\cite{Andriluka14CVPR}, to incorporate more appearance variation in the training process.
Half of each mini-batch is filled with examples of either kind.
Supervision via 2D labels is straightforward when using 2.5D heatmaps, as the $X$ and $Y$ heatmap axes correspond to the space in which the 2D labels are defined.
However, since our prediction volume is defined on a metric scale and is not aligned with image space, we propose a 2D loss computation method that is invariant to prediction scale and translation.
To this end, we first orthographically project the predicted 3D skeleton onto the image plane by discarding the Z coordinate.
Then we align the projected prediction to the 2D pixel-scale ground truth by translation and uniform scaling to the least-squares optimal fit before computing the loss.
This alignment layer is differentiable and gradients can be backpropagated through it.
We note that a similar scale-invariant loss has been used by Rhodin \etal to enforce multi-view consistency of 3D poses~\cite{Rhodin18CVPR}.
\subsection{Truncated Pose Estimation}
Our metric-space heatmap representation decouples the image boundary from the heatmap boundary.
This enables the prediction of joint locations outside the image frame without additional design effort, the network is simply trained to output complete poses at a metric scale, regardless of how the input image is scaled or cropped.
To evaluate this aspect, we follow Vosoughi \etal~\cite{Vosoughi18ICIP} by randomly cropping H3.6M inputs, keeping at least 1/4 of the area of the person bounding square.
Examples of such crops are in the second row of Fig.~\ref{fig:h36m-qualitative}.
We consider two scenarios.
In the first one, the above described sampling of truncated crops is only performed at test time.
In the second case, such crops are used for training as well.
\subsection{Training}
Prior work has shown that the $L^1$ loss is preferable in soft-argmax-based pose estimation~\cite{Sun18ECCV}.
To balance the losses computed on 3D and 2D-annotated examples, we use a fixed weighting factor $\lambda=0.1$ tuned on a separate validation set of Human3.6M, yielding the overall loss as

\begin{align}
\label{eq:loss-metro}
\mathcal{L} = \mathcal{L}_{\text{ann3D}} +  \lambda \mathcal{L}_{\text{ann2D}}.
\end{align}

We initialize the network with ImageNet-pretrained weights and use the Adam optimizer with weight decay~\cite{Loshchilov19ICLR} and a batch size of 64.
We decay the learning rate exponentially by an overall factor of 100, in two parts: from $10^{-4}$ to $3.33\times 10^{-5}$ over 25 epochs and from $3.33\times 10^{-6}$ to $10^{-6}$ in 2 final cooldown epochs.

As usual in deep learning, several sources of randomness influence the exact results of an experiment: random weight initialization, data shuffling, data augmentation and hardware-level non-determinism of execution order.
We control these (except the last) by consistently seeding the random number generators.
To distinguish random fluctuations from algorithmic differences, we repeat our experiments with 5 different seeds and report the mean and standard deviation of the evaluation metrics.

\begin{figure}[!t]
\centering
\includegraphics[width=\linewidth]{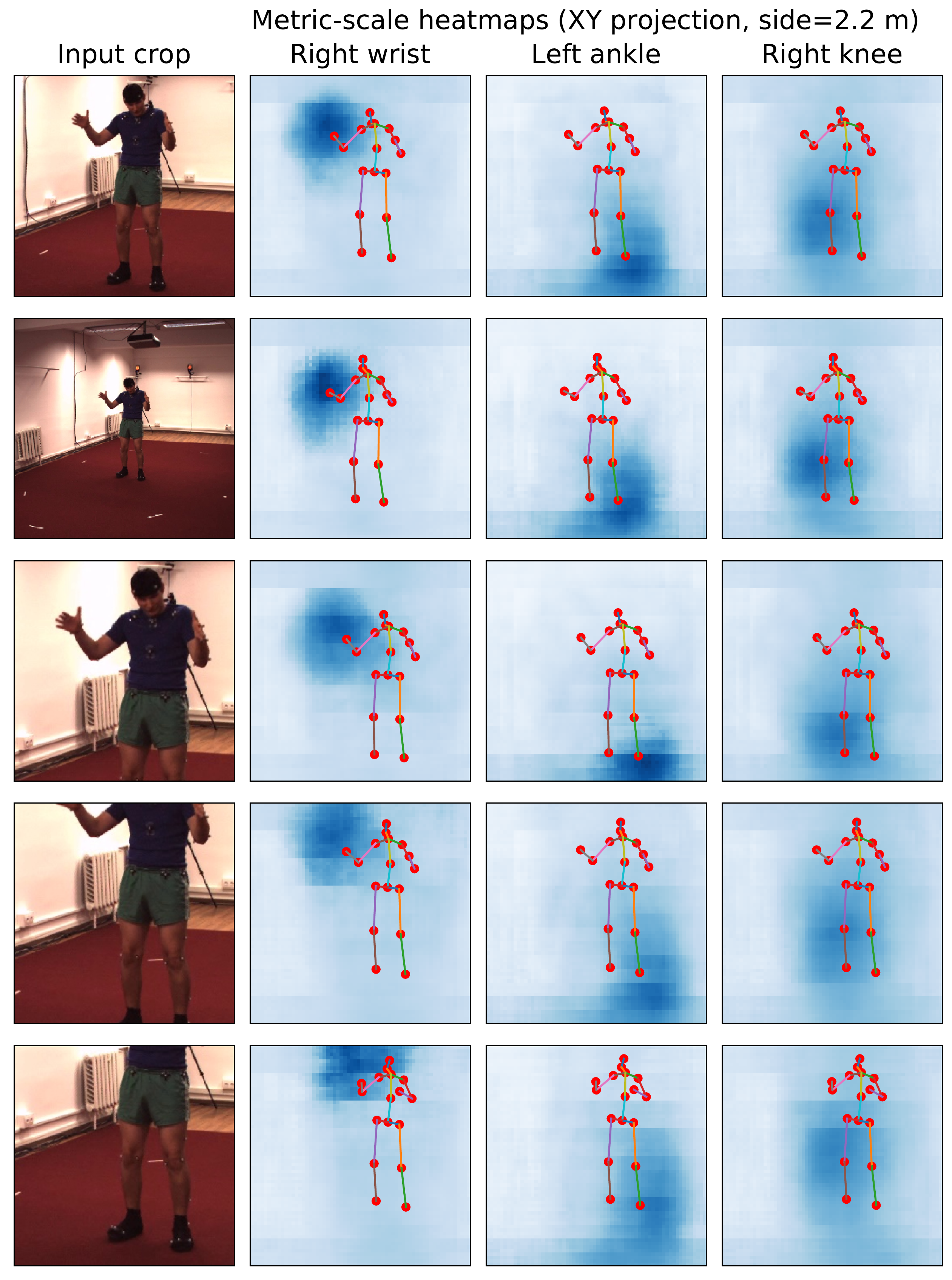}\\
\caption{A closer look at scale and truncation robustness.
We plot the projected \textit{metric-scale} heatmaps for 3 joints with the full soft-argmax skeleton for reference.
The predicted skeleton is approximately invariant to change in scale and truncation.
Since the metric size of the person does not change with image scaling, the backbone learns to output heatmaps with a similar center of mass, regardless of image scale.
Note that the heatmaps do not align with image space and this is intended by design.
(The broad peaks are a result of training the model at low, 8$\times$8 heatmap resolution.)
}
\label{fig:proj_heatmaps}
\end{figure}

\subsection{Intuition}
As described above, our network is trained to output complete skeletons at the same metric scale, regardless of image zooming and truncation.
To gain more intuition, we illustrate in Fig.~\ref{fig:proj_heatmaps} how this fully-convolutional model is able to achieve approximately image-scale- and truncation-invariant predictions.
In particular, we can see that the soft-argmax output is not necessarily in the middle of the heatmap's most prominent peak.
As soft-argmax yields the heatmap's center of mass, even distant heatmap values have an influence.
Intuitively, this allows the network to move the prediction result towards different heatmap locations by adding counter-balancing correction weights, for example at the image sides or at the person center.
Regarding truncation, the last row shows that the model can infer that the arms must lie above waist level, as there is no visual evidence of them in the image.
To understand how a fully-convolutional network can ``know'' where the truncation happens, we refer to Islam~\etal's paper~\cite{Islam20ICLR}, showing that even fully-convolutional networks can encode positional information as a result of the zero-paddings within convolutional layers.
This means that the location of the top image border can be used as a cue for the network to shift the full skeleton downwards inside the heatmap volume, such that it fits.
Note that the network is free to place the skeleton anywhere within the volume, since the root prediction is subtracted before computing the root-relative loss.
\section{Multi-Person Absolute Pose Approach}
\label{sec:absolute}

In this section, we propose MeTRAbs, a combination of MeTRo 3D heatmap estimation presented in Sec.~\ref{sec:metro} with traditional 2D pose heatmaps in a single end-to-end trained network for absolute 3D pose estimation.
The main idea is that the MeTRo approach implicitly estimates the scale, which we then use to infer the distance.
By applying this method within a top-down paradigm (detection, cropping, pose estimation), we obtain a fast and accurate way to tackle multi-person absolute 3D pose estimation.

Using our approach from Sec.~\ref{sec:metro}, we estimate a complete metric-scale pose $\left\{(\Delta X_j, \Delta Y_j, \Delta Z_j)^T\right\}_{j=1}^{J}$ up to translation (where $J$ is the number of joints).

By additionally estimating the 2D, image-space pose $\left\{(x_j, y_j)^T\right\}_{j=1}^{J}$, we obtain all the necessary information to recover the absolute 3D pose in the (calibrated) camera coordinate system, as we will see in the following.
For absolute pose estimation we assume known camera intrinsics, since monocular focal length estimation~\cite{Workman15ICIP,Kar15ICCV} is a very challenging task (\cf the ``dolly zoom'' effect~\cite{Liang20CVPRW}).
However, note that our method does not require the intrinsic calibration for root-relative estimation.

The absolute pose can be expressed as $\left\{(X_0 + \Delta X_j, Y_0 + \Delta Y_j, Z_0 + \Delta Z_j)^T\right\}_{j=1}^{J}$ with $(X_0, Y_0, Z_0)$ being the absolute pose offset, which we aim to recover.
For this, we first calculate the normalized image coordinates as $(\tilde{x}_j, \tilde{y}_j)^T = K^{-1}(x_j, y_j)^T$, where $K$ is the intrinsic matrix.

Mehta \etal~\cite{Mehta17TDV} derive a formula to reconstruct the absolute root position using the weak perspective projection model.
V\'eges \etal~\cite{Veges19IJCNN}, while still operating in the weak perspective model, note that an approximation step involved in Mehta \etal's algorithm leads to worse performance.
Motivated by this, we derive a reconstruction method under the full perspective pinhole camera model and extensively compare it with Mehta \etal's weak perspective method.
In a full perspective model, a perfect estimate would satisfy

\begin{equation}
\label{eq:pinhole}
\begin{bmatrix} \tilde{x}_j \\ \tilde{y}_j \end{bmatrix} = \begin{bmatrix} (X_0 + \Delta X_j)/(Z_0 + \Delta Z_j) \\ (Y_0 + \Delta Y_j)/(Z_0 + \Delta Z_j) \end{bmatrix}, \\
\end{equation}%
which can be rearranged to

\begin{equation}
\begin{bmatrix} X_0 - \tilde{x}_j Z_0  \\ Y_0 - \tilde{y}_j Z_0 \end{bmatrix} = \begin{bmatrix} \tilde{x}_j \Delta Z_j-\Delta X_j \\ \tilde{y}_j \Delta Z_j-\Delta Y_j \end{bmatrix}. \\
\end{equation}

Considering all joints, we obtain $2J$ linear equations in the three variables $(X_0, Y_0, Z_0)$.
Since $\tilde{x}, \tilde{y}, X, Y$ and $Z$ are estimates, the equation system is noisy and over-determined.
Hence we opt to solve it by linear least squares, with TensorFlow's differentiable solver based on Cholesky decomposition.
This differentiability allows us to directly supervise the network with a loss $\mathcal{L}^{\text{abs3D}}$ computed on the final absolute 3D pose output, which turns out to be crucial for accurate distance estimation.

For truncated images, Eq.~\ref{eq:pinhole} only holds for body joints inside the image frame, since the 2D heatmap method cannot estimate out-of-image joint locations.
We therefore exclude joints from the optimization, which are predicted to lie closer to the image border than one stride length.
After reconstructing the root joint position, we can obtain the absolute pose in two ways.
Either as $(\Delta X_j+X_0, \Delta Y_j+Y_0, \Delta Z_j+Z_0)^T$ (adding the reconstructed offset to the 3D head's root-relative output), or as $(\tilde{x}_j, \tilde{y}_j, 1)^T\cdot(\Delta Z_j+Z_0)$ (back-projecting the 2D head's output).
For joints that lie within the image, we use the latter option, while for truncated ones we use the former.
Both the individual prediction heads and the final absolute output are supervised with the $L^1$ loss.
As in the root-relative MeTRo network, we apply weak supervision from 2D-labeled data for MeTRAbs as well, on both heads.
Extending Eq.~\ref{eq:loss-metro}, the loss becomes

\begin{align}
\begin{split}
\mathcal{L} =& \mathcal{L}_{\text{ann3D}}^{\text{abs3D}} + \mathcal{L}_{\text{ann3D}}^{\text{head3D}} + \mathcal{L}_{\text{ann3D}}^{\text{head2D}}
+ \lambda \left(\mathcal{L}_{\text{ann2D}}^{\text{head2D}} + \mathcal{L}_{\text{ann2D}}^{\text{head3D}}\right),
\end{split}
\end{align}
where we again set $\lambda = 0.1$.

We found that the absolute loss can introduce numerical instabilities very early during training, since at this point the two prediction heads do not yet produce sufficiently compatible outputs, making the reconstruction problem ill-conditioned.
Hence, we only turn on the absolute loss after 5000 update steps.

In a multi-person scenario, inference speed becomes a priority, since the model is evaluated on each person detection separately.
To retain real-time performance, we do not apply dense prediction with MeTRAbs; the network is trained and tested with coarse, 8$\times$8 heatmaps.
\section{2.5D Baseline}

\begin{figure}[t]
\centering
\includegraphics[width=\columnwidth]{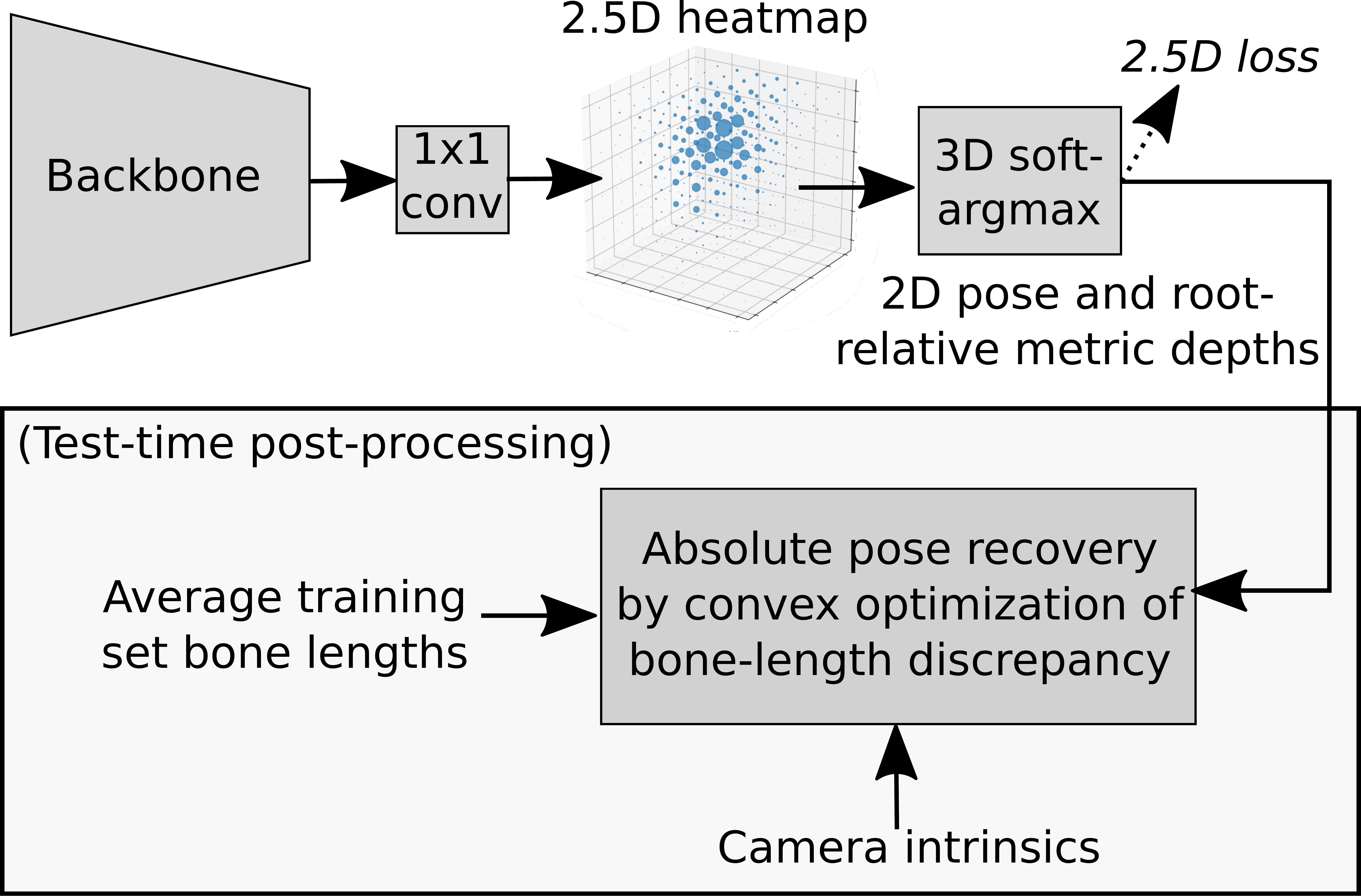} \\
\caption{Baseline architecture with 2.5D heatmaps for ablative comparison experiments.}
\label{fig:baseline}
\end{figure}

\label{sec:baseline}
For comparison, we implement a 2.5D baseline derived from Pavlakos~\etal's work~\cite{Pavlakos17CVPR}, which introduced volumetric heatmaps for 3D human pose estimation.
Pavlakos \etal~use a coarse-to-fine estimation scheme with a stacked hourglass architecture~\cite{Newell16ECCV} and no soft-argmax.
To make the baseline directly comparable to our results, we instead use the architecture depicted in Fig.~\ref{fig:baseline}.
This baseline directly estimates 2.5D heatmaps through a 1$\times$1 convolution at the end of the backbone.
We then use soft-argmax, and compute the $L^1$ loss on the resulting coordinates.
This makes the baseline similar to Sun \etal~\cite{Sun18ECCV}, except Sun \etal used additional learned layers and did not perform scale recovery.
As a test-time post-processing step, the baseline uses the bone-length optimization method from Pavlakos~\etal~\cite{Pavlakos17CVPRsupp} to recover the root joint depth, which we briefly reiterate here.
Given an assumed value for the root joint depth $Z_0$ and known camera intrinsics, the 2.5D pose can be back-projected into metric space and each bone's resulting length $b_i(Z_0)$ can be calculated.
The optimal $Z_0$ is then the one that minimizes the squared bone length discrepancy, as compared to the average training bone lengths $t_{i}$:

\begin{align}
\begin{split}
Z_0^* &= \arg\min_{Z_0} \sum_{i\in\text{bones}}\left(b_{i}(Z_0) - t_{i}\right)^2, \\
\end{split}
\label{eq:baseline}
\end{align}
where we only use bones, whose both ends are predicted to lie within the image (further from the border than 1 stride length).
This is a convex, nonlinear least-squares problem, and we solve it using the Levenberg-Marquardt algorithm initialized at $Z_0=2\text{ m}$.
To reiterate, as in~\cite{Pavlakos17CVPR}, the absolute pose is not supervised during the baseline's training and the convex optimization of $Z_0$ is not backpropagated through, for simplicity.
We note, however, that the recent development of differentiable convex optimization layers~\cite{Amos17ICML,Agrawal19NeurIPS} could, in principle, enable such a solution as well.
\section{Datasets, Preprocessing, Evaluation}
We conduct our single-person experiments on the largest available benchmarks: Human3.6M (shortened as H3.6M)~\cite{Ionescu11ICCV,Ionescu14PAMI} and MPI-INF-3DHP (3DHP)~\cite{Mehta17TDV}.
The extended approach described in Sec.~\ref{sec:absolute} is evaluated in a multi-person context by training on MuCo-3DHP (MuCo) and testing on MuPoTS-3D (MuPoTS).

\textbf{H3.6M}~\cite{Ionescu11ICCV,Ionescu14PAMI} was captured with 4 cameras in a motion capture studio.
Two evaluation protocols are in wide use.
In Protocol 1, the training subjects are 1, 5, 6, 7, 8, while 9 and 11 are used for testing.
Prediction and ground-truth are aligned at the root joint, but no Procrustes alignment is performed.
In Protocol 2, subjects 1, 5, 6, 7, 8, 9 are used in training and 11 in evaluation, with Procrustes alignment between prediction and ground truth.
Every 64\textsuperscript{th} frame is evaluated.
We use the provided bounding boxes.
We downsample videos from 50 to 10 fps.
To further reduce redundancy, training frames are only used if at least one body joint moves at least 100 mm since the previous kept frame.

\textbf{MPII}~\cite{Andriluka14CVPR} is a 2D-labeled dataset with 25k training images.
We use this dataset for weak supervision, following the idea of Zhou~\etal\cite{Zhou17ICCV}.
Only arm and leg joints are used from MPII, as we found these to be the most consistently labeled across datasets.
In single-person experiments we only use instances explicitly marked as ``well-separated'' from other people and take the provided person centers and sizes as the center and side length of the bounding box.
In multi-person experiments, we use all person instances and the boxes are obtained with YOLOv3~\cite{Redmon18Arxiv}.

\textbf{3DHP}~\cite{Mehta17TDV} shows 8 training subjects in a green-screen studio.
Test frames come from 3 scenes, each with 2 subjects: green-screen studio, studio without green screen, and outdoor.
The latter two make this benchmark more challenging than H3.6M.
In this dataset, the hips are labeled closer to the legs than in MPII.
Following \cite{Zhou17ICCV}, we move the hips towards the neck by a fifth of the pelvis-neck vector before comparing with MPII-annotated skeletons for 2D loss computation.
3DHP provides two ground truth variants: unnormalized metric-space poses and ``universal'' (height-normalized) ones.
We evaluate on both.
We use only the chest-height cameras as~\cite{Mehta17TDV}, and only examples where all joints are within the image.
We generate 3DHP bounding boxes by combining the bounding box of labeled joints and the most confident person detection of YOLOv3.
The same frame sampling strategy is used as described above for H3.6M.
Since its publication, the official 3DHP ground truth has been changed twice, making not all published results comparable.
In our experience the first update changes scores by 1-3\%, while the second one only by 0.1\%, which is within experimental fluctuation, making the two latest versions comparable.

\textbf{MuCo}~\cite{Mehta18TDV} is a synthetically composited multi-person dataset, derived from 3DHP by pasting persons over each other based on their root joint depth order.
As \cite{Veges19IJCNN}, we generate 150k training images, each with 4 people.
We run YOLOv3 on these images to get realistic bounding boxes.

\textbf{MuPoTS}~\cite{Mehta18TDV} is a mixed indoor and outdoor multi-person test set, compatible with MuCo, consisting of 20 sequences showing people performing various actions and interactions.
Like 3DHP, MuPoTS also provides normalized and unnormalized skeletons.

\begin{table*}[t]
\caption{Evaluation on H3.6M Protocol 1 (subjects 9 and 11), using mean per joint position error (MPJPE) without Procrustes alignment.\newline All methods use extra 2D-labeled pose data in training.}
\setlength\tabcolsep{2.0mm}
\centering
\begin{tabu}{@{}lrrrrrrrrrrrrrrrc@{}}
\toprule
& \centercell{Dir.} & \centercell{Dis.}&\centercell{Eat}  & \centercell{Gre.} & \centercell{Phn.} & \centercell{Pose}  & \centercell{Pur.}& \centercell{Sit}   & \centercell{SitD}  & \centercell{Sm.} & \centercell{Pht.} & \centercell{Wait}  & \centercell{Walk} & \centercell{WD} &\multicolumn{1}{c}{WT} & Avg $\downarrow$ \\
\midrule
\multicolumn{17}{c}{\textit{Methods using ground-truth scale or depth information at test time}} \\
\midrule 
Sun \smalletal~\cite{Sun17ICCV}                  & 52.8 & 54.8 & 54.2 & 54.3 & 61.8 & 53.1 & 53.6 & 71.7 & 86.7 & 61.5 & 67.2 & 53.4 & 47.1 & 61.6 & 53.4 & 59.1 \\
Nibali \smalletal~\cite{Nibali19WACV} &--&--&--&--&--&--&--&--&--&--&--&--&--&--&--&  57.0 \\
Luvizon \smalletal~\cite{Luvizon18CVPR}          & 51.5 & 53.4 & 49.0 & 52.5 & 53.9 & 50.3 & 54.4 & 63.6 & 73.5 & 55.3 & 61.9 & 50.1 & \bf46.0 & 60.2 & 51.0 & 55.1 \\
Luvizon \smalletal~\cite{Luvizon20PAMI} & 43.7 & 48.8 & 45.6 & \bf46.2 & 49.3 & 43.5 & 46.0 & 56.8 & 67.8 & 50.5 & 57.9 & \bf43.4 & 40.5 & 53.2 & 45.6 & 49.5 \\
Sun \smalletal~\cite{Sun18ECCV}        & 47.5 & 47.7 & 49.5 & 50.2 & \un{51.4} & 43.8 & \un{46.4} & \un{58.9} & 65.7 & \un{49.4} & \un{55.8} & \un{47.8} & 38.9 & {\bf49.0} & \un{43.8} & \un{49.6} \\
Chen \smalletal~\cite{Chen19BMVC}       & 45.3 & 49.8 & 46.1 & 49.6 & {\bf48.2} & {\bf41.7} & 47.4 & {\bf53.1} & {\bf55.2} & {\bf48.0} & 57.7 & 45.6 & 40.8 & 52.4 & 45.2 & {\bf48.4} \\ 
\midrule
\multicolumn{17}{c}{\textit{Methods using no ground truth scale or depth information at test time}} \\
\midrule
Pavlakos \smalletal~\cite{Pavlakos17CVPR}        & 67.4 & 72.0 & 66.7 & 69.1 & 72.0 & 77.0 & 65.0 & 68.3 & 83.7 & 96.5 & 71.7 & 65.8 & 74.9 & 59.1 & 63.2 & 71.9 \\
Zhou \smalletal~\cite{Zhou17ICCV}                & 54.8 & 60.7 & 58.2 & 71.4 & 62.0 & 53.8 & 55.6 & 75.2 & 111.6& 64.2 & 65.5 & 66.0 & 51.4 & 63.2 & 55.3 & 64.9 \\
Martinez \smalletal~\cite{Martinez17ICCV}        & 51.8 & 56.2 & 58.1 & 59.0 & 69.5 & 55.2 & 58.1 & 74.0 & 94.6 & 62.3 & 78.4 & 59.1 & 49.5 & 65.1 & 52.4 & 62.9 \\
Fang \smalletal~\cite{Fang18AAAI} 	            & 50.1 & 54.3 & 57.0 & 57.1 & 66.6 & 53.4 & 55.7 & 72.8 & 88.6 & 60.3 & 73.3 & 57.7 & 47.5 & 62.7 & 50.6 & 60.4 \\
Yang \smalletal~\cite{Yang18CVPR}                & 51.5 & 58.9 & 50.4 & 57.0 & 62.1 & 49.8 & 52.7 & 69.2 & 85.2 & 57.4 & 65.4 & 58.4 & 43.6 & 60.1 & 47.7 & 58.6 \\
Pavlakos \smalletal~\cite{Pavlakos18CVPR}        & 48.5 & 54.4 & 54.4 & 52.0 & 59.4 & 49.9 & 52.9 & 65.8 & 71.1 & 56.6 & 65.3 & 52.9 & 44.7 & 60.9 & 47.8 & 56.2 \\
Liu \smalletal~\cite{Liu19WACV}              & \un{47.0} & 53.1 & 50.3 & \un{48.8} & 56.0 & 48.1 & 47.6 & 65.9 & 72.6 & 52.3 & 61.4 & 49.1 & 39.3 & 54.2 & {\bf40.6} & 52.4 \\
Xu \smalletal~\cite{Xu20CVPR} & \bf40.6 & \bf47.1 & 45.7 & \bf46.6 & 50.7 & 45.0 & 47.7 & 56.3 & \bf63.9 & 49.4 & 63.1 & \bf46.5 & \bf38.1 & 51.9 & 42.3 & \bf49.2 \\
Sharma \smalletal~\cite{Sharma19ICCV}  & 48.6 & 54.5 & 54.2 & 55.7 & 62.6 & 50.5 & 54.3 & 70.0 & 78.3 & 58.1 & 72.0 & 55.4 & 45.2 & 61.4 & 49.7 & 58.0 \\
Cai \smalletal~\cite{Cai19ICCV}& 46.5 & 48.8 & 47.6 & 50.9 & 52.9 & 48.3 & 45.8 & 59.2 & 64.4 & 51.2 & 61.3 & 48.4 & 39.2 & 53.5 & 41.2 & 50.6 \\
\midrule
2.5D baseline & 45.1 & 50.4 & 45.4 & 47.8 & \bf50.0 & \bf44.6 & 49.8 & 59.0 & 69.4 & 49.4 & 56.5 & 48.0 & 39.6 & \bf49.4 & 45.0 & 50.2\std{0.3} \\
\textbf{MeTRo} (ours) & 46.3 & 48.3 & \bf43.3 & 48.2 & 50.2 & 45.1 & \bf46.1 & \bf56.2 & 66.8 & \bf49.3 & \bf54.5 & 46.7 & 40.1 & 49.6 & 46.2 & {\bf49.3\std{0.7}} \\
\bottomrule \\
\end{tabu}
\label{tab:h36m_protocol1}
\end{table*}

\begin{table*}[t]
\caption{Comparison of MPJPE with prior work on H3.6M under Protocol 2 (test subject 11 with Procrustes alignment to the ground truth). }
\setlength\tabcolsep{0.8mm}
\centering
\begin{tabular}{@{}cccccccccccc@{}}
\toprule
 & Nie \cite{Nie17ICCV} & Pavlakos \cite{Pavlakos17CVPR} & Sun \cite{Sun17ICCV} & Martinez \cite{Martinez17ICCV} & Sun \cite{Sun18ECCV} & Nibali \cite{Nibali19WACV} & Habibie \cite{Habibie19CVPR} & Xu \cite{Xu20CVPR}& Chen \cite{Chen19BMVC} & 2.5D baseline  & MeTRo (ours) \\
\midrule
P-MPJPE & 79.5 & 51.9 & 48.3 & 47.7 & 40.6 & 40.4 & 49.2 & 38.9 & {\bf 33.7} & 34.5\std{0.4} & 34.7\std{0.5} \\
\bottomrule
\end{tabular}
\vspace{-2mm}
\label{tab:h36m_other_protocol}
\end{table*}%

We crop images to the person's bounding square and resize it to $256\times 256$ px.
Perspective effects must be taken into account when centering the image on the subject as this induces an implicit camera rotation~\cite{Mehta17TDV}.
We compensate by transforming both the input image and the predicted pose according to the implied camera rotation.
The indoor 3DHP and MuPoTS sequences are gamma-corrected with an exponent of 0.67.

We apply geometric \textbf{augmentations} (scaling, rotation, translation, horizontal flip) and color distortion (brightness, contrast, hue, saturation).
For single-person datasets, synthetic occlusion is added with 70\% probability, half of which are rectangles with uniform white noise as in~\cite{Zhong17AAAI}, half are segmented non-person objects from Pascal VOC~\cite{Everingham12} as in~\cite{Sarandi18IROSW,Sarandi18Arxiv}.
On MuCo, synthetic occlusion probability is reduced to 30\% since some occlusion is already introduced from compositing person segments over each other.
On 3DHP and MuCo, we also augment the background with 70\% probability following~\cite{Mehta17TDV}.
Backgrounds are taken from INRIA Holidays~\cite{Jegou08ECCV}, excluding person images.
All evaluation is done on a single crop, with no test-time augmentation.

We use the standard \textbf{evaluation measures}.
The main one on 3DHP and MuPoTS is the percentage of correct keypoints (PCK), \ie the fraction of joints predicted within 150 mm of the ground truth.
The AUC is the area under the PCK curve as the threshold ranges from 0 to 150 mm.
The measure on H3.6M is the mean per joint position error (MPJPE).
3DHP and MuPoTS evaluate 14 joints, excluding the root, while H3.6M uses 17, including the root.
The official MuPoTS evaluation script rescales the bone lengths of the prediction to match the ground truth bone lengths before computing metrics, leading to some confusion and inconsistency between reported results.
In \cite{Mehta18TDV} rescaling was only used for evaluating LCR-Net~\cite{Rogez17CVPR}, but it has since been adopted by other authors as well.
For consistency and simplicity, we train MeTRAbs only with unnormalized skeletons.
When evaluating on universal (normalized) skeletons, we use bone rescaling.
On unnormalized skeletons, we do not use bone rescaling, in order to directly evaluate the raw metric-space outputs of the methods.
Following V\'eges \etal~\cite{Veges20Arxiv}, on MuPoTS we also evaluate absolute (\ie non-root-relative) versions of these metrics, prefixed with ``A-'', \eg A-PCK.
For absolute MPJPE, V\'eges \etal~\cite{Veges19IJCNN,Veges20Arxiv} evaluate all 17 joints and 16 (no pelvis) for relative MPJPE (but use 14 for PCK and A-PCK).
For consistency, we always use 14 joints on MuPoTS, except when marked otherwise.
\section{Results}
\label{sec:results}

\begin{table*}[t]
\caption{Comparison on MPI-INF-3DHP with prior methods. 
\st{}Evaluated with the first version of the dataset, with some annotation difference. Dashes (--) reflect a lack of published information. Superscripts indicate the training data (first characters of 3DHP, H36M, MPII, LSP and COCO).}
\setlength\tabcolsep{1.7mm}
\centering
\begin{tabu}{lccccccc|ccc|ccc}
\toprule
& \makecell{Stand/\\ walk} &\makecell{Exer-\\ cise} &\makecell{Sit on\\ chair}&\makecell{Cro./\\ reach}&\makecell{On\\ floor}&\makecell{Sport}&\makecell{Misc.}&\makecell{Green\\screen}&\makecell{No\\gr.sc.}&\makecell{Out-\\ door}&\multicolumn{3}{c}{Total} \\
\midrule
 & \multicolumn{10}{c|}{PCK$\uparrow$} & PCK$\uparrow$ & AUC$\uparrow$ & MPJPE$\downarrow$ \\
\midrule
\multicolumn{14}{c}{\textit{Universal, height-normalized skeletons (simplified scale recovery task)}} \\
\midrule 
Rogez \smalletal~\cite{Rogez17CVPR}\st & 70.5 & 56.3 & 58.5 & 69.4 & 39.6 & 57.7 & 57.6 & -- & -- & -- & 59.7 & 27.6 & 158.4 \\
Zhou \smalletal\textsuperscript{H+M}~\cite{Zhou17ICCV}\st & 85.4 & 71.0 & 60.7 & 71.4 & 37.8 & 70.9 & 74.4 & 71.7 & 64.7 & 72.7 & 69.2 & 32.5 & 137.1 \\
Zhou \smalletal\textsuperscript{H+M}~\cite{Zhou19ICCV} & -- & -- & -- & -- & -- & -- & -- & 75.6 & 71.3 & 80.3 & 75.3 & 38.0 &  -- \\
Mehta \smalletal\textsuperscript{3+M+L+H}~\cite{Mehta17TOG}\st & 87.7 & 77.4 & 74.7 & 72.9 & 51.3 & 83.3 & 80.1 & -- & -- & -- & 76.6 & 40.4 & 124.7 \\ 
Mehta \smalletal\textsuperscript{3+M+L+H}~\cite{Mehta17TDV}\st & 86.6 & 75.3 & 74.8 & 73.7 & 52.2 & 82.1 & 77.5 & 84.6 & 72.4 & 69.7 & 75.7 & 39.3 & 117.6 \\
Mehta \smalletal\textsuperscript{3+M+L+C}~\cite{Mehta18TDV}\st & 83.8 & 75.0 & 77.8 & 77.5 & 55.1 & 80.4 & 72.5 & -- & -- & -- & 75.2 & 37.8 & 122.2 \\
Luo \smalletal\textsuperscript{3+M+H}~\cite{Luo18BMVC,Luo18Github} & {\bf95.5} & 82.3 & 89.9 & 84.6 & 66.5 & 92.0 & 93.0 & -- & -- & -- & 84.3 & 47.5 & 84.5 \\
Nibali \smalletal\textsuperscript{3+M}~\cite{Nibali19WACV} & -- & -- & -- & -- & -- & -- & -- & -- & -- & -- & 87.6 & 48.8 &  87.6 \\
\midrule
2.5D baseline\textsuperscript{3+M} & 95.1 & 90.7 & 86.8 & {\bf92.4} & {\bf74.2} & 94.1 & 91.7 & 92.1 & 89.0 & {\bf87.7} & 89.9\std{0.2} & 52.8\std{0.4} & 79.7\std{0.6} \\
\textbf{MeTRo} (ours)\textsuperscript{3+M} & 95.0 & {\bf91.8} & {\bf90.2} & 92.1 & 73.4 & {\bf95.1} & {\bf91.8} & {\bf93.4} & {\bf90.3} & 86.5 & {\bf90.6}\std{0.4} & {\bf56.2}\std{0.5} & {\bf74.9}\std{1.4} \\
\midrule
\multicolumn{14}{c}{\textit{Metric-scale skeletons (full scale recovery task)}} \\
\midrule
2.5D baseline\textsuperscript{3+M} & 93.1 & {\bf89.3} & 83.6 & {\bf93.1} & {\bf73.7} & {\bf93.2} & {\bf91.1} & 89.0 & {\bf87.9} & {\bf89.4} & {\bf88.7}\std{0.6} & 48.6\std{1.3} & {\bf87.1}\std{2.2} \\
\textbf{MeTRo} (ours)\textsuperscript{3+M} & {\bf94.0} & 89.2 & {\bf87.1} & 89.1 & 68.9 & 92.6 & 90.3 & {\bf90.1} & 87.8 & 85.7 & 88.2\std{0.5} & {\bf48.7}\std{0.7} & 88.4\std{1.3} \\
\bottomrule
\end{tabu}
\label{tab:3dhp}
\end{table*}

\begin{table}[t]
\caption{Comparison with baseline methods of scale recovery, with or without access to ground truth information. For both datasets, metric-scale skeletons are used with the same 17 joints for comparability. The first two comparison methods access the ground truth at test time.}
\centering
\setlength\tabcolsep{0.7mm}
\begin{tabu}{@{}lccc|ccc@{}}
\toprule
 & \multicolumn{3}{c|}{H3.6M} & \multicolumn{3}{c}{3DHP}\\
 & PCK$\uparrow$ & AUC$\uparrow$ & MPJPE$\downarrow$ & PCK$\uparrow$ & AUC$\uparrow$ & MPJPE$\downarrow$ \\
\midrule
2.5D GT root depth    & 96.6 & \it68.8 &\it49.0 & \it90.8 & \it56.1 & \it74.2  \\
2.5D GT bone length   & 96.4 & 67.0 & 51.9& 90.3 & \it56.1 & 74.6 \\
\midrule
2.5D avg train bones & 96.6  & 68.1 & 50.2 & \bf89.6 & 52.1 & \bf80.6 \\
\textbf{MeTRo} (ours) & \bf97.0 & \bf68.6 & \bf49.3 & \bf89.6 & \bf52.6 & 81.1  \\
\bottomrule
\end{tabu}
\vspace{-2mm}
\label{tab:baselines}
\end{table}%

\begin{table}[t]
\vspace{0pt}
\centering
\setlength\tabcolsep{0.5mm}
\caption{MPJPE scores on H3.6M under truncation, evaluating all or only the present joints. (\st{}Training was not performed with truncated crops.) Results of other methods are taken from ~\cite{Vosoughi18ICIP}.}
\begin{tabularx}{\linewidth}{@{}lccccc@{}}
\toprule
 & Mehta\st\cite{Mehta17TOG} & Zhou\st\cite{Zhou17ICCV} & Vosoughi\cite{Vosoughi18ICIP} & {\bf MeTRo}\st & {\bf MeTRo}  \\
 \midrule
All joints & 396.4 & 400.5 & 185.0 & 124.7 & \textbf{77.8} \\
Present joints & 338.0  & 332.5 & 173.6 & 76.8 & \textbf{59.8} \\
\bottomrule
\end{tabularx}
\label{tab:h36m_partial_presence}
\end{table}%

\begin{table}[t]
\caption{Test speed (crops per second, FPS) and error (H3.6M MPJPE) tradeoff with the two striding variants from Fig.~\ref{fig:striding}.}
\centering
\begin{tabular}{@{}llcccc@{}}
\toprule
& \multirow{2}{50pt}{\makecell{Striding \\ variant} } & \multicolumn{4}{c}{Test stride} \\
 \cmidrule{3-6}
 & &  32 & 16 & 8 & 4 \\
\midrule
\multirow{2}{30pt}{MPJPE}& normal strides & 53.1 & 52.5 & 52.7 & 52.9 \\
& center-aligned & 50.9 & 50.2 & 50.0 & {\bf 49.3}  \\
\midrule
\multirow{2}{50pt}{\makecell[l]{Speed \\ (crop per sec.)} } & no batching & 160 & 150 & 105 & 38 \\
& batch size 8 & {\bf 511} & 475 & 292 & 92 \\
\bottomrule
\end{tabular}
\vspace{-2mm}
\label{tab:strides}
\end{table}%

\begin{table}[t]
\caption{Augmentation ablation on H36M.}
\centering
\begin{tabular}{@{}cccc@{}}
\toprule
Geometry & Color & Occlusion  & MPJPE \\
\midrule
\checkmark & -- & -- & 58.0 \\
\checkmark & \checkmark & -- & 52.8 \\
\checkmark & \checkmark & \checkmark & {\bf 49.3} \\
\bottomrule
\end{tabular}
\label{tab:augmentation}
\end{table}

\begin{table}[!t]
\centering
\caption{Results on MuPoTS-3D. Detected, unnormalized poses, no bone rescaling. (\st{}Re-evaluated public results; joint count: $^\dagger$17, $^\ddagger$16, else 14)
}
\setlength\tabcolsep{0.9pt}
\begin{tabularx}{\linewidth}{lccccc}
\toprule
 & A-MPJPE$\downarrow$ & MPJPE$\downarrow$ &  A-PCK$\uparrow$  & PCK$\uparrow$ & Det.Rate$\uparrow$   \\
\midrule
Rogez \smalletal \cite{Rogez17CVPR} & -- & 146$^\ddagger$ & -- & -- & 86 \\
Mehta \smalletal \cite{Mehta18TDV} & -- & 132$^\ddagger$  & -- & -- & 93 \\ 
Baseline in \cite{Veges19IJCNN} & 320$^\dagger$ & 122$^\ddagger$  & -- & -- & 91 \\
V\'eges \smalletal \cite{Veges19IJCNN} & 292$^\dagger$ & 120$^\ddagger$  & -- & -- & 91 \\
V\'eges \smalletal \cite{Veges20Arxiv}\st & 257.2 (255$^\dagger$) & 119.4 (108$^\ddagger$) & 38.1 & 75.4 & 93 \\ 
\midrule
2.5D baseline & 317.6 (313.6$^\dagger$) & 114.0 (110.0$^\ddagger$) & \bf40.0\std{1.0} & 79.3\std{0.3} & {\bf94.2}\std{0.0} \\
{\bf MeTRAbs} & {\bf248.2} ({\bf246.9}$^\dagger$) & {\bf108.2} ({\bf104.3}$^\ddagger$) & \bf40.2\std{1.9} & {\bf81.1}\std{0.4} & 94.1\std{0.1}\\ 
\quad w/o abs. loss & 328.8 (327.8$^\dagger$) & 108.4 (104.7$^\ddagger$) & 36.7\std{3.2} & 80.9\std{0.4} & 94.1\std{0.1} \\
\bottomrule
\end{tabularx}
\label{tab:mupots_abs}
\end{table}

\begin{table}[!t]
\centering
\caption{Comparison of weak (W) and full (F) perspective-based absolute pose reconstruction. The two letters denote the training and the test time variant. (Unnormalized skeletons, without bone rescaling.)
}
\setlength\tabcolsep{7pt}
\begin{tabularx}{\linewidth}{cccccc}
\toprule
\multicolumn{2}{c}{Persp. assumption} & \multicolumn{2}{c}{All annotations}& \multicolumn{2}{c}{Matched annotations} \\
\midrule
Training & Test & A-PCK$\uparrow$  & PCK$\uparrow$ & A-PCK$\uparrow$ & PCK$\uparrow$ \\
\midrule
F  &  F  & 37.2\std{1.7} & 76.2\std{0.5} &39.3\std{1.7} & 79.9\std{0.5}\\ 
F  &  W  & {\bf39.4}\std{1.6} & 76.2\std{0.5} &{\bf41.6}\std{1.6}&80.0\std{0.5}\\ 
W  &  F  & 35.6\std{1.8} & 77.1\std{0.4} &37.6\std{1.8}&81.0\std{0.5}\\ 
W  &  W  & 38.1\std{1.8} & {\bf77.2}\std{0.4} &40.2\std{1.9}&{\bf81.1}\std{0.4}\\ 
-- & F &  33.0\std{3.3} & 77.0\std{0.4} &34.9\std{3.3}&80.8\std{0.4}\\
-- & W & 34.8\std{3.1} & 77.0\std{0.4} &36.7\std{3.2}&80.9\std{0.4}\\
\bottomrule
\end{tabularx}
\label{tab:mupots_persp}
\end{table}
\begin{table}[t]
\caption{Results on the 3DPW challenge dataset. (PA=Procrustes analysis)}
\centering
\setlength\tabcolsep{0.6mm}
\begin{tabularx}{\linewidth}{lcccc}
\toprule
& MPJPE$\downarrow$ & MPJPE-PA$\downarrow$ & PCK@50mm$\uparrow$ & AUC@200mm$\uparrow$  \\
\midrule
\multicolumn{5}{c}{\textit{Known association to GT identity}} \\
\midrule
Sun~\smalletal~\cite{Sun20Arxiv} & 81.8 & 58.6 & 37.3 & 59.9 \\
Kissos~\smalletal~\cite{Kissos20Arxiv} & 83.2 & 59.7 & 42.4 & 62.3 \\
MeTRAbs (ours) & \bf68.8 & \bf49.7 & \bf48.8 & \bf66.8 \\
\midrule
\multicolumn{5}{c}{\textit{Unknown association to GT identity}} \\
\midrule
MeTRAbs (ours) & 85.1 & 56.7 & 45.8 & 63.2 \\
\bottomrule
\end{tabularx}
\label{tab:tdpw}
\end{table}%

\begin{table*}[!t]
\centering
\setlength\tabcolsep{3.5pt}
\caption{Comparison to prior work on the MuPoTS-3D benchmark for normalized skeletons with bone rescaling to the ground truth before computing the percentage of correct keypoints (PCK).
(For the direct evaluation of the metric-space poses, see Tab. \ref{tab:mupots_abs}).}
\begin{tabularx}{\linewidth}{lccccccccccccccccccccc}
\toprule
 & S1 & S2 & S3 & S4 & S5 & S6 & S7 & S8 & S9 & S10 & S11 & S12 & S13 & S14 & S15 & S16 & S17 & S18 & S19 & S20 & Avg$\uparrow$ \\
\midrule
\multicolumn{22}{c}{\textit{Root-relative PCK for all annotated poses}} \\
\midrule
Rogez \smalletal~\cite{Rogez17CVPR} & 67.7 & 49.8 & 53.4 & 59.1 & 67.5 & 22.8 & 43.7 & 49.9 & 31.1 & 78.1 & 50.2 & 51.0 & 51.6 & 49.3 & 56.2 & 66.5 & 65.2 & 62.9 & 66.1 & 59.1 & 53.8 \\
Mehta \smalletal~\cite{Mehta18TDV} & 81.0 & 60.9 & 64.4 & 63.0 & 69.1 & 30.3 & 65.0 & 59.6 & 64.1 & 83.9 & 68.0 & 68.6 & 62.3 & 59.2 & 70.1 & 80.0 & 79.6 & 67.3 & 66.6 & 67.2 & 66.0 \\
Rogez \smalletal~\cite{Rogez19PAMI} & 87.3 & 61.9 & 67.9 & 74.6 & 78.8 & 48.9 & 58.3 & 59.7 & 78.1 & 89.5 & 69.2 & 73.8 & 66.2 & 56.0 & 74.1 & 82.1 & 78.1 & 72.6 & 73.1 & 61.0 & 70.6 \\
Moon \smalletal~\cite{Moon19ICCV} & \bf94.4 & 77.5 & 79.0 & 81.9 & 85.3 & 72.8 & 81.9 & \bf75.7 & \bf90.2 & 90.4 & 79.2 & 79.9 & 75.1 & \bf72.7 & 81.1 & 89.9 & 89.6 & 81.8 & \bf81.7 & \bf76.2 & 81.8 \\
Dabral \smalletal~\cite{Dabral19TDV} & 85.1 & 67.9 & 73.5 & 76.2 & 74.9 & 52.5 & 65.7 & 63.6 & 56.3 & 77.8 & 76.4 & 70.1 & 65.3 & 51.7 & 69.5 & 87.0 & 82.1 & 80.3 & 78.5 & 70.7 & 71.3 \\
V\'eges \smalletal~\cite{Veges20Arxiv} & 89.5 & 75.9 & 85.2 & 83.9 & 85.0 & 73.4 & 83.6 & 58.7 & 65.1 & 90.4 & 76.8 & 81.9 & 67.0 & 55.9 & 80.8 & 90.6 & \bf90.0 & 81.1 & 81.1 & 68.6 & 78.2 \\
Mehta \smalletal~\cite{Mehta19Arxiv} & 89.7 & 65.4 & 67.8 & 73.3 & 77.4 & 47.8 & 67.4 & 63.1 & 78.1 & 85.1 & 75.6 & 73.1 & 65.0 & 59.2 & 74.1 & 84.6 & 87.8 & 73.0 & 78.1 & 71.2 & 72.1 \\
Benzine~\smalletal~\cite{Benzine20PR} & 78.1 & 62.5 & 55.5 & 63.8 & 70.2 & 50.8 & 73.8 & 65.3 & 55.1 & 79.3 & 70.4 & 72.3 & 65.4 & 55.3 & 65.2 & 81.3 & 77.2 & 75.9 & 74.2 & 71.6 & 67.5 \\
\midrule
2.5D baseline & 93.0 & 76.4 & 88.6 & 85.2 & 86.3 & 75.7 & 84.3 & 67.9 & 84.3 & \bf93.4 & 81.6 & 89.8 & 77.3 & 67.7 & 83.8 & 91.0 & 86.1 & 84.8 & 77.1 & 71.2 & 82.3\std{0.1} \\
\textbf{MeTRAbs} & 93.8 & 80.8 & \bf89.3 & \bf87.0 & 86.6 & 74.5 & 83.7 & 66.2 & 85.0 & 92.9 & 80.4 & 89.6 & 77.1 & 68.7 & \bf86.3 & 92.0 & 86.6 & 84.4 & 77.3 & 71.4 & 82.7\std{0.3} \\
\quad{} w/o abs. loss & 94.0 & \bf82.6 & 88.4 & 86.5 & \bf87.3 & \bf76.2 & \bf85.9 & 66.9 & 85.8 & 92.9 & \bf81.8 & \bf89.9 & \bf77.6 & 68.5 & 85.6 & \bf92.3 & 89.3 & \bf85.1 & 78.2 & 71.6 & {\bf83.3}\std{0.2} \\
\midrule
\multicolumn{22}{c}{\textit{Root-relative PCK for detected poses}}  \\
\midrule
Rogez \smalletal~\cite{Rogez17CVPR} & 69.1 & 67.3 & 54.6 & 61.7 & 74.5 & 25.2 & 48.4 & 63.3 & 69.0 & 78.1 & 53.8 & 52.2 & 60.5 & 60.9 & 59.1 & 70.5 & 76.0 & 70.0 & 77.1 & 81.4 & 62.4 \\
Mehta \smalletal~\cite{Mehta18TDV} & 81.0 & 65.3 & 64.6 & 63.9 & 75.0 & 30.3 & 65.1 & 61.1 & 64.1 & 83.9 & 72.4 & 69.9 & 71.0 & 72.9 & 71.3 & 83.6 & 79.6 & 73.5 & 78.9 & 90.9 & 70.8 \\
Rogez \smalletal~\cite{Rogez19PAMI} & 88.0 & 73.3 & 67.9 & 74.6 & 81.8 & 50.1 & 60.6 & 60.8 & 78.2 & 89.5 & 70.8 & 74.4 & 72.8 & 64.5 & 74.2 & 84.9 & 85.2 & 78.4 & 75.8 & 74.4 & 74.0 \\
Moon \smalletal~\cite{Moon19ICCV} & \bf94.4 & 78.6 & 79.0 & 82.1 & 86.6 & 72.8 & 81.9 & 75.8 & \bf90.2 & 90.4 & 79.4 & 79.9 & 75.3 & 81.0 & 81.0 & 90.7 & 89.6 & 83.1 & 81.7 & 77.3 & 82.5 \\
Dabral \smalletal~\cite{Dabral19TDV} & 85.8 & 73.6 & 61.1 & 55.7 & 77.9 & 53.3 & 75.1 & 65.5 & 54.2 & 81.3 & 82.2 & 71.0 & 70.1 & 67.7 & 69.9 & 90.5 & 85.7 & 86.3 & 85.0 & 91.4 & 74.2 \\
V\'eges \smalletal~\cite{Veges20Arxiv} & 89.5 & 81.6 & 85.9 & 84.4 & 90.5 & 73.5 & 85.5 & 68.9 & 65.1 & 90.4 & 79.1 & 82.6 & 72.7 & 68.1 & 81.0 & 94.0 & \bf90.4 & 87.4 & \bf90.4 & 92.6 & 82.7 \\
Mehta \smalletal~\cite{Mehta19Arxiv} & 89.7 & 78.6 & 68.4 & 74.3 & 83.7 & 47.9 & 67.4 & 75.4 & 78.1 & 85.1 & 78.7 & 73.8 & 73.9 & 77.9 & 74.8 & 87.1 & 88.3 & 79.5 & 88.3 & \bf97.5 & 78.0 \\
Benzine~\smalletal~\cite{Benzine20PR} & 78.3 & 75.0 & 56.9 & 64.1 & 76.1 & 51.3 & 74.7 & \bf79.1 & 55.2 & 79.3 & 74.5 & 74.5 & 70.2 & 69.5 & 67.6 & 85.7 & 82.6 & 78.7 & 79.1 & 89.6 & 72.7 \\
\midrule
2.5D baseline & 93.0 & 80.1 & 89.2 & 85.8 & 90.1 & 76.9 & 88.6 & 75.6 & 84.3 & \bf93.4 & 85.9 & 90.6 & 83.4 & 80.9 & 83.8 & 93.0 & 86.6 & 89.3 & 85.0 & 90.8 & 86.3\std{0.1} \\
\textbf{MeTRAbs} & 93.8 & 84.4 & \bf90.0 & \bf87.6 & 90.5 & 75.7 & 88.1 & 74.9 & 85.0 & 92.9 & 84.7 & 90.4 & 83.3 & \bf82.2 & \bf86.3 & 93.9 & 87.1 & 88.9 & 85.2 & 91.3 & 86.8\std{0.4} \\
\quad{} w/o abs. loss & 94.0 & \bf86.5 & 89.0 & 87.1 & \bf91.1 & \bf77.4 & \bf90.2 & 75.7 & 85.8 & 92.9 & \bf86.0 & \bf90.7 & \bf83.8 & 82.0 & 85.6 & \bf94.3 & 89.8 & \bf89.6 & 86.5 & 91.7 & {\bf87.5}\std{0.2} \\
\midrule
\multicolumn{22}{c}{\textit{Absolute PCK for all annotated poses}}  \\
\midrule
Moon~\smalletal~\cite{Moon19ICCV} & 59.5 & 44.7 & 51.4 & 46.0 & 52.2 & 27.4 & 23.7 & 26.4 & \bf39.1 & 23.6 & 18.3 & 14.9 & 38.2 & 26.5 & 36.8 & 23.4 & 14.4 & 19.7 & 18.8 & 25.1 & 31.5 \\
Benzine~\smalletal~\cite{Benzine20PR} & 22.2 & 18.1 & 16.1 & 18.5 & 20.4 & 14.7 & 21.2 & 18.9 & 16.0 & 22.9 & 20.3 & 20.9 & 18.9 & 16.0 & 18.9 & 23.5 & 22.3 & 21.8 & 21.5 & 20.8 & 19.8 \\
V\'eges~\smalletal\cite{Veges20Arxiv} & 50.4 & 33.4 & 52.8 & 27.5 & 53.7 & 31.4 & 22.6 & 33.5 & 38.3 & 56.5 & 24.4 & 35.5 & 45.5 & 34.9 & 49.3 & 23.2 & \bf32.0 & 30.7 & 26.3 & \bf43.8 & 37.3 \\
\midrule
2.5D baseline & \bf77.6 & \bf50.5 & \bf58.6 & 40.3 & \bf74.6 & 21.9 & 7.3 & 27.0 & 22.4 & 38.6 & \bf32.2 & 37.6 & 25.2 & 43.9 & 50.4 & \bf35.0 & 25.5 & \bf41.1 & \bf31.9 & 27.8 & \bf38.5\std{0.9} \\
\textbf{MeTRAbs} & 21.2 & 21.1 & 45.5 & \bf48.2 & 40.9 & \bf34.9 & \bf33.0 & \bf51.5 & 34.9 & \bf85.6 & 18.0 & 36.7 & \bf50.3 & \bf53.1 & \bf54.3 & 28.1 & 28.8 & 26.8 & 20.0 & 35.1 & \bf38.4\std{1.9} \\
\quad{} w/o abs. loss & 48.9 & 32.9 & 15.3 & 18.9 & 48.7 & 11.8 & 19.1 & 42.3 & 28.9 & 78.4 & 27.5 & \bf60.6 & 38.6 & 42.8 & 43.1 & 28.4 & 28.7 & 28.6 & 23.3 & 33.8 & 35.0\std{3.1} \\
\midrule
\multicolumn{22}{c}{\textit{Absolute PCK for detected poses}} \\
\midrule

Moon~\smalletal~\cite{Moon19ICCV} & 59.5 & 45.3 & \bf51.4 & 46.2 & 53.0 & 27.4 & 23.7 & 26.4 & \bf39.1 & 23.6 & 18.3 & 14.9 & 38.2 & 29.5 & 36.8 & 23.6 & 14.4 & 20.0 & 18.8 & 25.4 & 31.8 \\
Benzine~\smalletal~\cite{Benzine20PR} & 22.7 & 21.2 & 17.1 & 18.6 & 22.0 & 14.8 & 21.5 & 22.9 & 16.0 & 22.9 & 21.5 & 21.6 & 20.3 & 20.0 & 19.4 & 18.9 & 23.8 & 22.6 & 22.9 & 25.8 & 20.9 \\
V\'eges~\smalletal\cite{Veges20Arxiv} & 50.4 & 35.9 & 53.3 & 27.7 & 57.2 & 31.4 & 23.1 & 39.3 & 38.3 & 56.5 & 25.2 & 35.8 & 49.3 & 42.5 & 49.4 & 24.1 & \bf32.1 & 33.1 & 29.3 & \bf59.2 & 39.6 \\
\midrule
2.5D baseline & \bf77.6 & \bf53.0 & \bf59.1 & 40.5 & \bf77.9 & 22.2 & 7.6 & 30.1 & 22.4 & 38.6 & \bf33.9 & 37.9 & 27.2 & 52.4 & 50.4 & \bf35.8 & 25.7 & \bf43.3 & \bf35.2 & 35.5 & \bf40.3\std{1.0} \\
\textbf{MeTRAbs} & 21.2 & 22.1 & 45.8 & \bf48.5 & 42.8 & \bf35.4 & \bf34.8 & \bf58.3 & 34.9 & \bf85.6 & 19.0 & 37.0 & \bf54.3 & \bf63.6 & \bf54.3 & 28.8 & 29.0 & 28.2 & 22.0 & 44.9 & \bf40.5\std{1.9} \\
\quad{} w/o abs. loss & 48.9 & 34.5 & 15.5 & 19.0 & 50.8 & 12.0 & 20.1 & 48.0 & 28.9 & 78.4 & 29.0 & \bf61.1 & 41.7 & 51.2 & 43.1 & 29.0 & 28.8 & 30.1 & 25.8 & 43.3 & 36.9\std{3.1} \\
\bottomrule
\end{tabularx}
\label{tab:mupots}
\end{table*}

\begin{figure}[t]
\centering
\includegraphics[height=65mm]{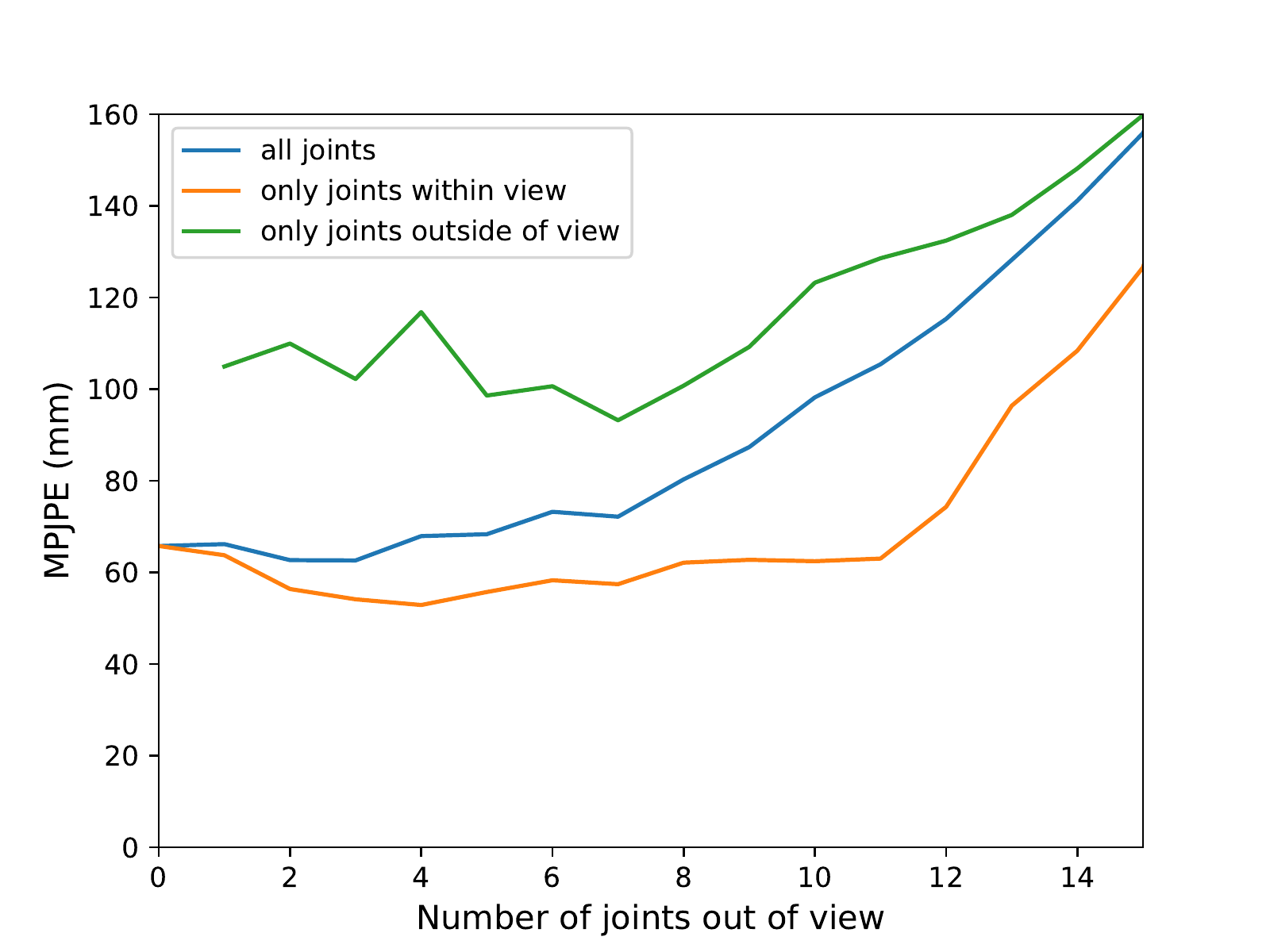} \\
\caption{Analysis of robustness to truncation on H3.6M.
Average performance remains relatively stable up to 7 truncated joints.}
\label{fig:fov}
\end{figure}

On \textbf{H3.6M} without ground truth depth or scale information, we achieve 49.3 mm MPJPE, which is within the margin of error compared to the state-of-the-art by Xu \etal~\cite{Xu20CVPR} (49.2 mm), while using a considerably simpler approach (see Tab.~\ref{tab:h36m_protocol1}).
(In all tables, the number after ``$\pm$'' is the standard deviation of 5 repeated experiments with different random seeds, therefore the standard error of the mean is a fifth of this value.)
Besides simplifying the prediction pipeline and allowing for truncation-robust prediction (see below), our metric heatmap representation also performs better than the 2.5D baseline with bone-length-based scale recovery under the exact same experimental conditions.
Tab.~\ref{tab:augmentation} shows that training data augmentations improve performance by a large margin.
On Protocol 2 (Tab.~\ref{tab:h36m_other_protocol}), the benefit of our method is masked by the use of Procrustes alignment, which explicitly ignores the quality of scale recovery.
It is therefore unsurprising that our method performs about equally well as the 2.5D variant.%

\begin{figure*}
\centering
\newlength{\imageheight}
\newlength{\imagegap}
\newlength{\labelsize}
\setlength{\labelsize}{14mm}
\setlength{\imageheight}{27mm}
\setlength{\imagegap}{0.8mm}
\makecell[{{p{\labelsize}}}]{H36M}\includegraphics[align=c,height=\imageheight]{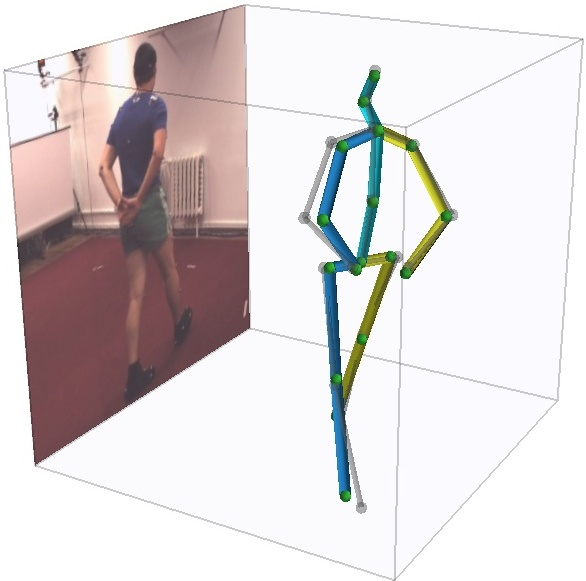}\hspace{\imagegap}%
\includegraphics[align=c,height=\imageheight]{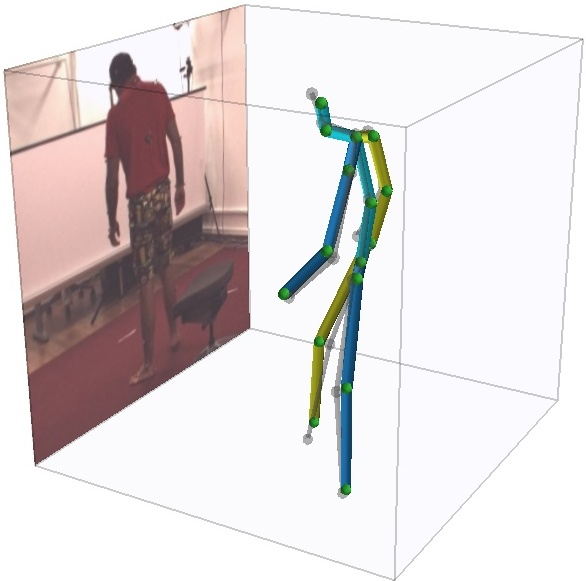}\hspace{\imagegap}%
\includegraphics[align=c,height=\imageheight]{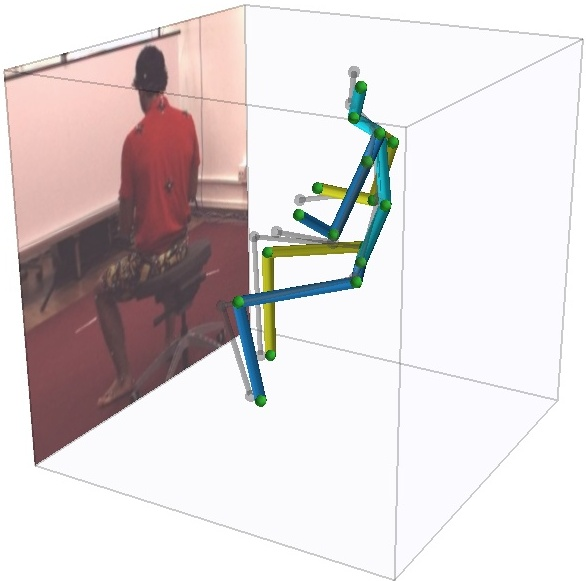}\hspace{\imagegap}%
\includegraphics[align=c,height=\imageheight]{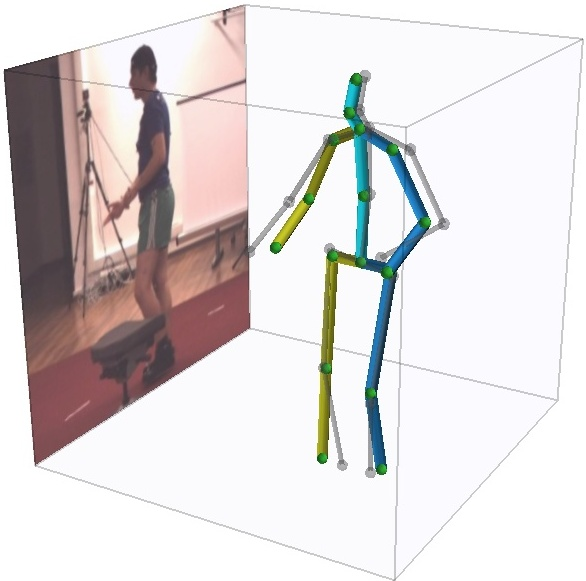}\hspace{\imagegap}%
\includegraphics[align=c,height=\imageheight]{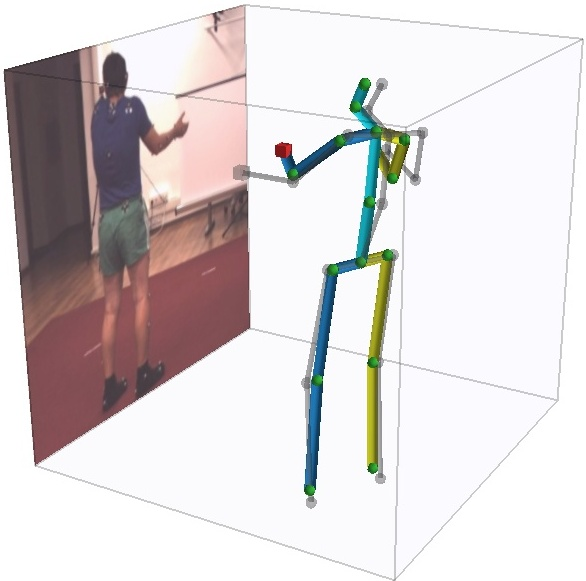}\hspace{-2.5mm}%
\includegraphics[align=c,height=7.7mm]{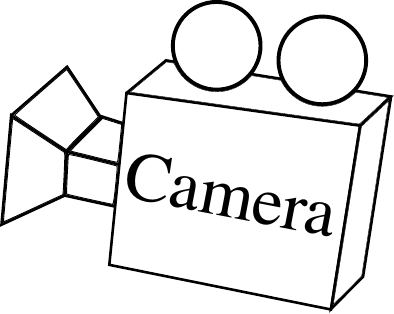}\\
\makecell[{{p{\labelsize}}}]{H3.6M \\ (partial body)}\includegraphics[align=c,height=\imageheight]{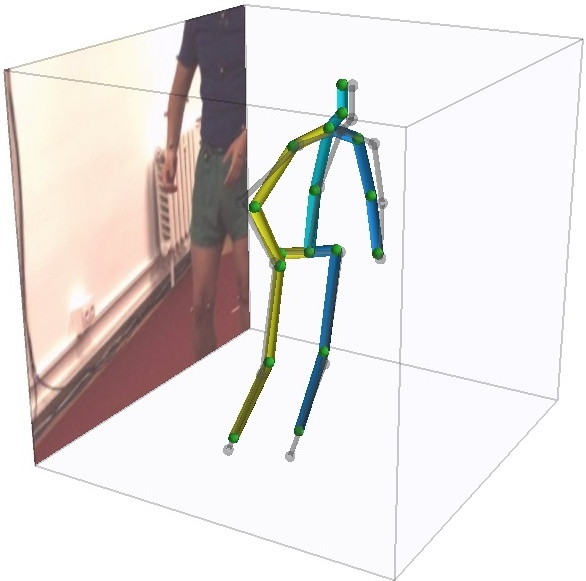}\hspace{\imagegap}%
\includegraphics[align=c,height=\imageheight]{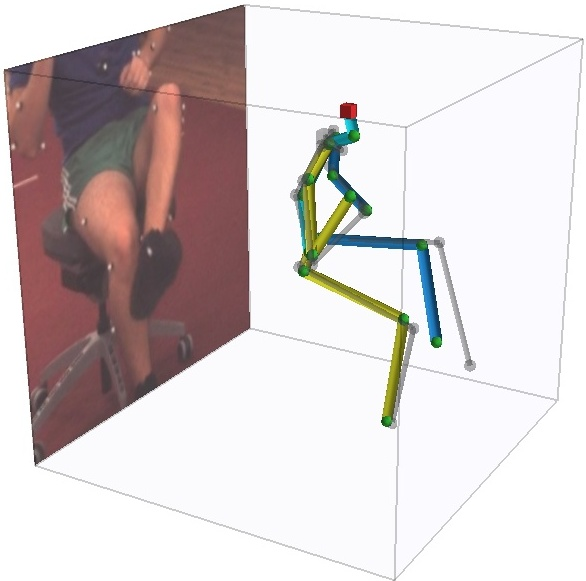}\hspace{\imagegap}%
\includegraphics[align=c,height=\imageheight]{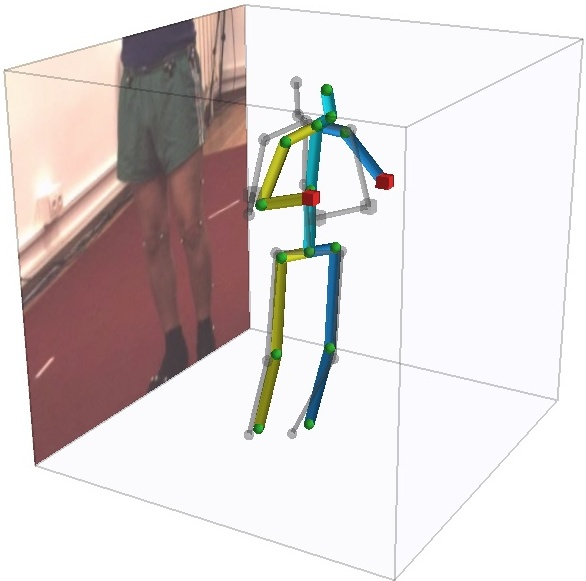}\hspace{\imagegap}%
\includegraphics[align=c,height=\imageheight]{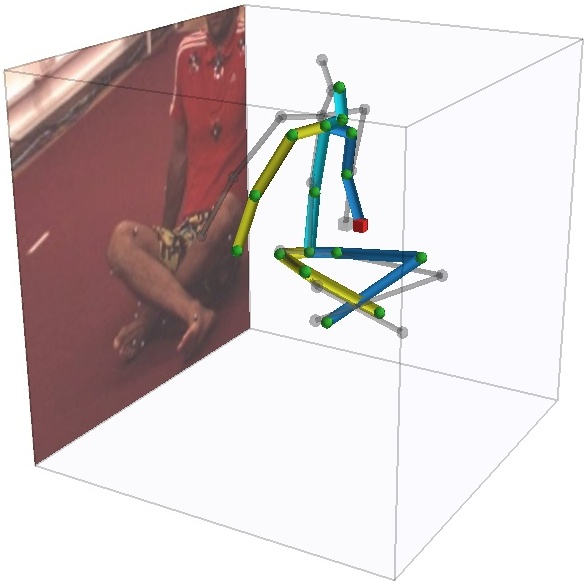}\hspace{\imagegap}%
\includegraphics[align=c,height=\imageheight]{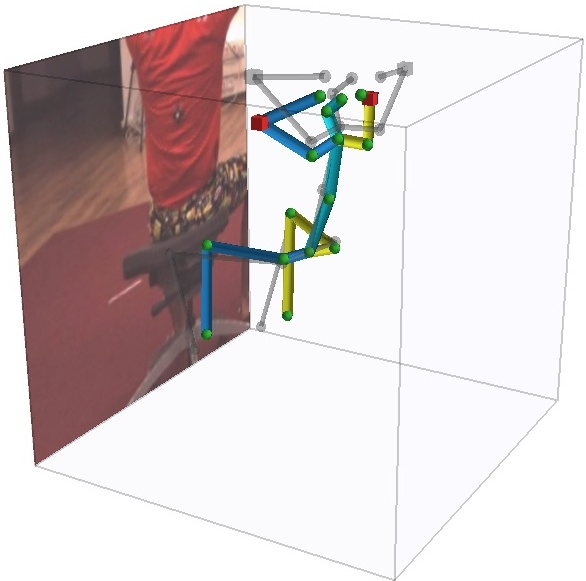}\hspace{-2.5mm}%
\includegraphics[align=c,height=7.7mm]{images/camera}\\ 	
\makecell[{{p{\labelsize}}}]{3DHP}\includegraphics[align=c,height=\imageheight]{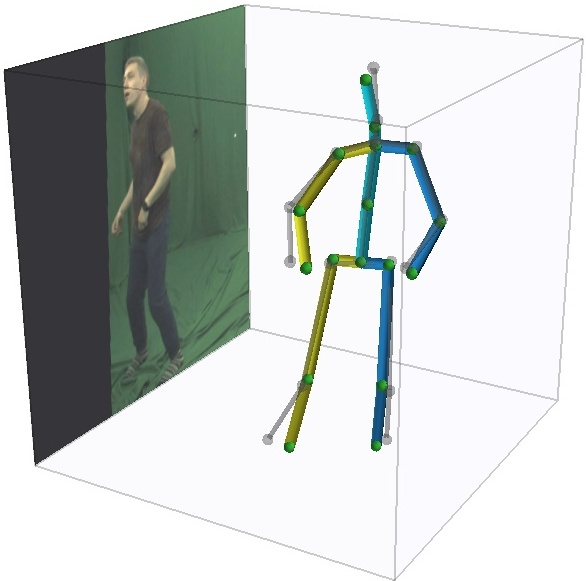}\hspace{\imagegap}%
\includegraphics[align=c,height=\imageheight]{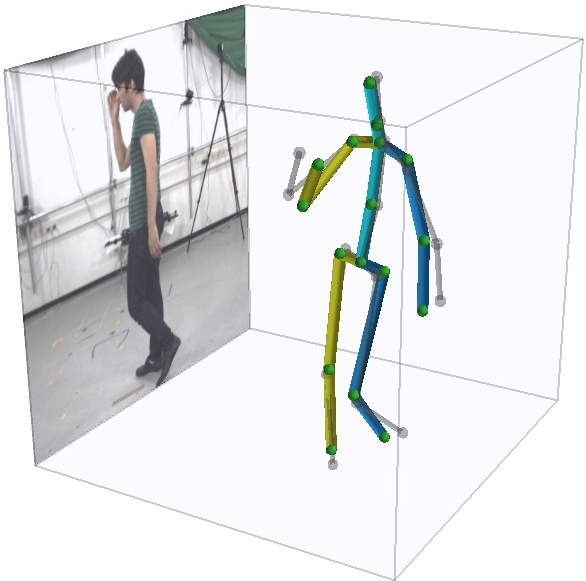}\hspace{\imagegap}%
\includegraphics[align=c,height=\imageheight]{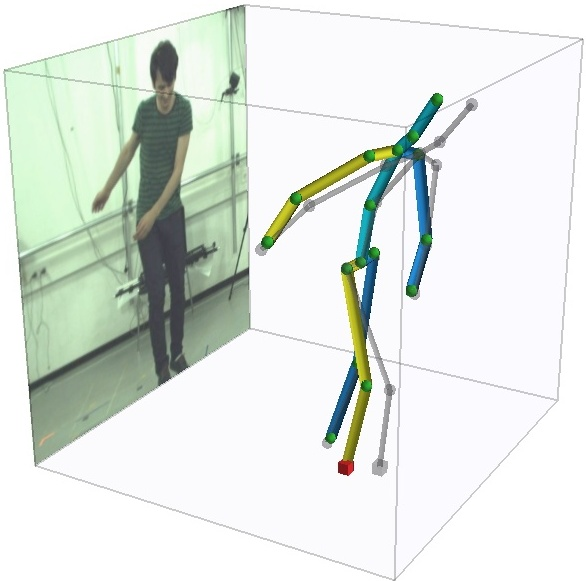}\hspace{\imagegap}%
\includegraphics[align=c,height=\imageheight]{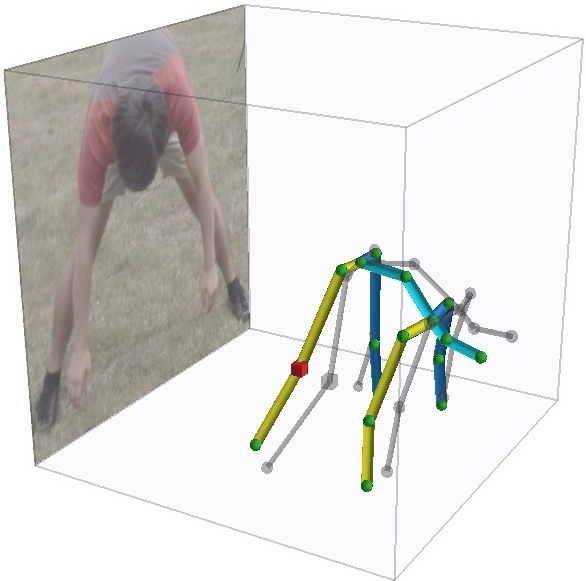}\hspace{\imagegap}%
\includegraphics[align=c,height=\imageheight]{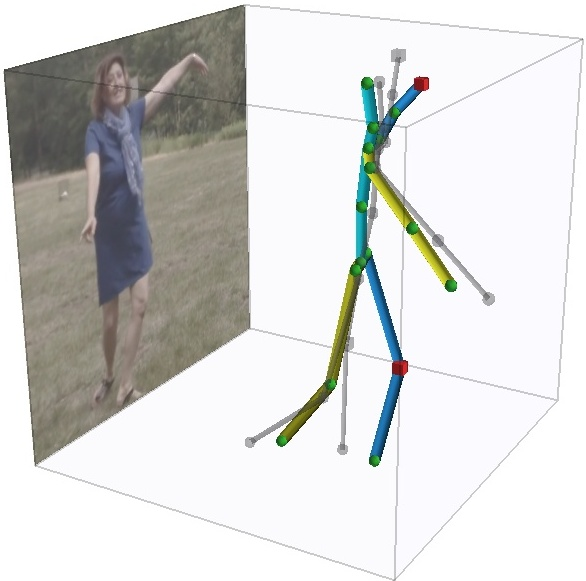}\hspace{-2.5mm}%
\includegraphics[align=c,height=7.7mm]{images/camera}\\
\makecell[{{p{\labelsize}}}]{MPII}\includegraphics[align=c,height=\imageheight]{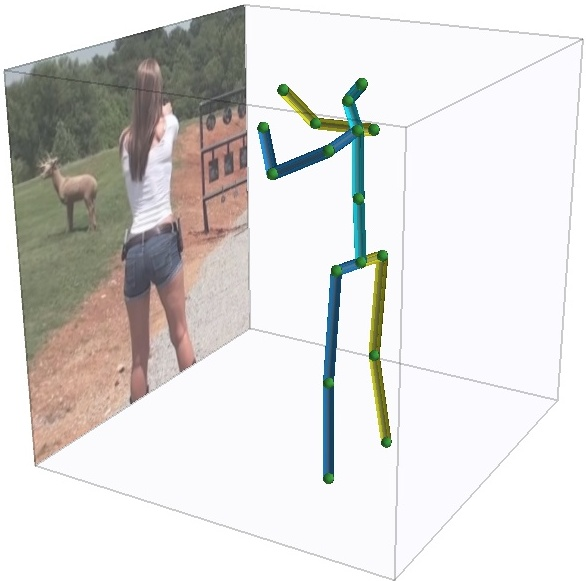}\hspace{\imagegap}%
\includegraphics[align=c,height=\imageheight]{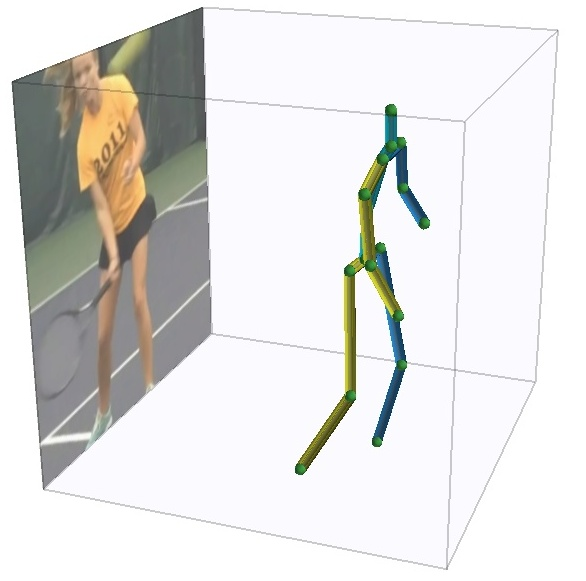}\hspace{\imagegap}%
\includegraphics[align=c,height=\imageheight]{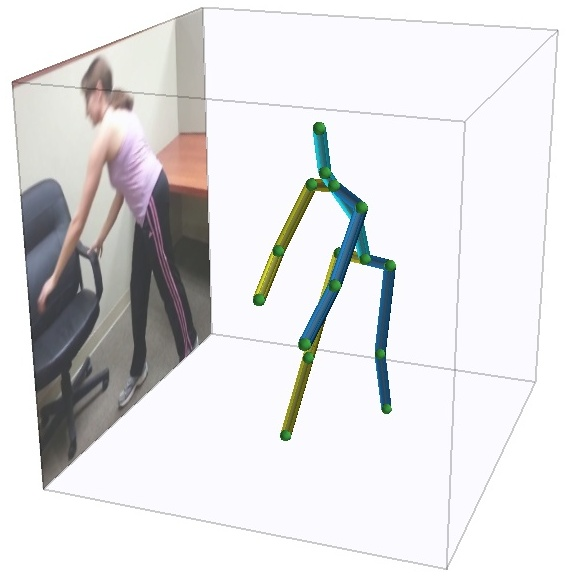}\hspace{\imagegap}%
\includegraphics[align=c,height=\imageheight]{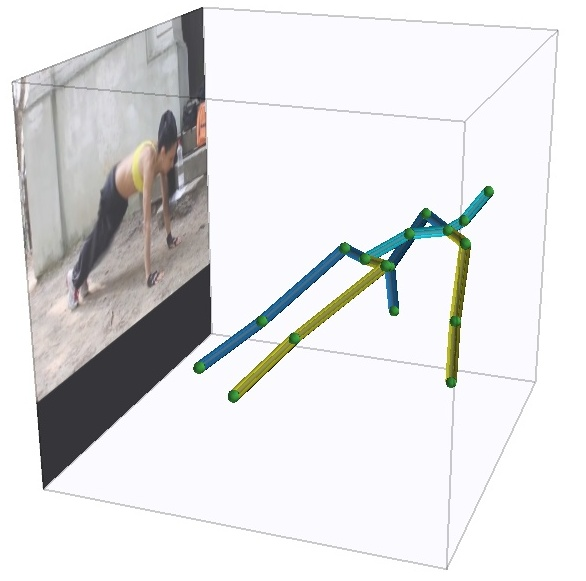}\hspace{\imagegap}%
\includegraphics[align=c,height=\imageheight]{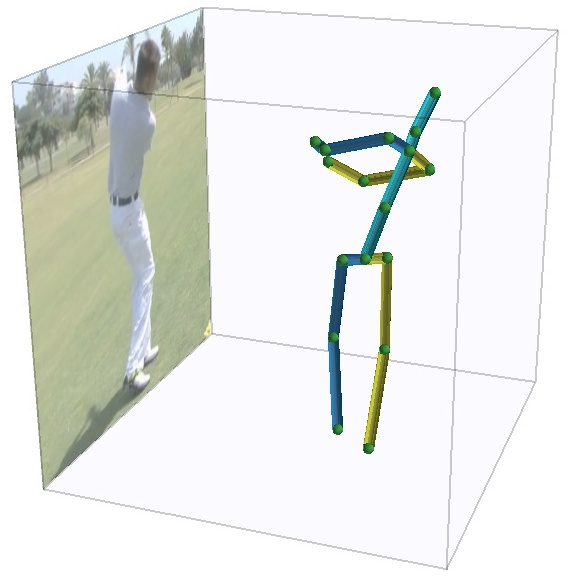}\hspace{-2.5mm}%
\includegraphics[align=c,height=7.7mm]{images/camera}\\
\vspace{-3mm}
\caption{Qualitative results on various datasets. Predictions are shown in color, ground truth in gray (except for MPII, where it is unavailable). Green spheres mark predictions within 150 mm of the ground truth, red cubes beyond that threshold. Note that our method performs well on truncated (partial body) images as well (second row).}
\label{fig:h36m-qualitative}
\end{figure*}

On \textbf{3DHP}, our method outperforms prior work by a large margin, including ones trained on more datasets as well (Tab.~\ref{tab:3dhp}).
Both with universal (height-normalized) skeletons and true metric-scale ones, the MeTRo representation outperforms the baseline on green-screen studio images, however, the outdoor scenes were recorded on an empty field without scale cues and the explicit bone-length-based scale recovery performs better there.
Qualitative results are in Fig.~\ref{fig:h36m-qualitative}.

We analyze \textbf{scale recovery} in more detail (Tab.~\ref{tab:baselines}).
As expected, the idealized method with test-time access to the ground truth root joint depth performs best on both H3.6M and 3DHP.
The proposed approach performs better than the 2.5D baseline using average bone lengths on H3.6M and comparably on 3DHP.
On H3.6M, MeTRo closes most of this scale recovery gap between the 2.5D average bone length baseline and the idealized variant using the true root.
Interestingly, our approach outperforms even the 2.5D variant using ground truth bone lengths for each test frame.
On 3DHP, MeTRo's scale recovery performance is similar to the 2.5D baseline (equal PCK, better AUC, slightly worse MPJPE).
Further, on this dataset, access to ground truth scale information provides a larger improvement than on H3.6M, highlighting the importance of testing on many subjects.

When tested on \textbf{truncated crops}, our method by far outperforms prior approaches (Tab.~\ref{tab:h36m_partial_presence}).
This is true even for our default training configuration, but performance improves substantially when training on truncated images as well.
The method is robust to truncation of up to 7 or 8 joints (of the 17) before overall performance substantially degrades (Fig.~\ref{fig:fov}).
Given the obvious ambiguity introduced by truncation, it is noteworthy that even truncated joints can be estimated with as little as about 100 mm average error.
Qualitative examples are in the second row of Fig.~\ref{fig:h36m-qualitative}, showing that our method can handle strongly truncated cases as well.

\begin{figure}[t]
\centering
\includegraphics[width=\columnwidth]{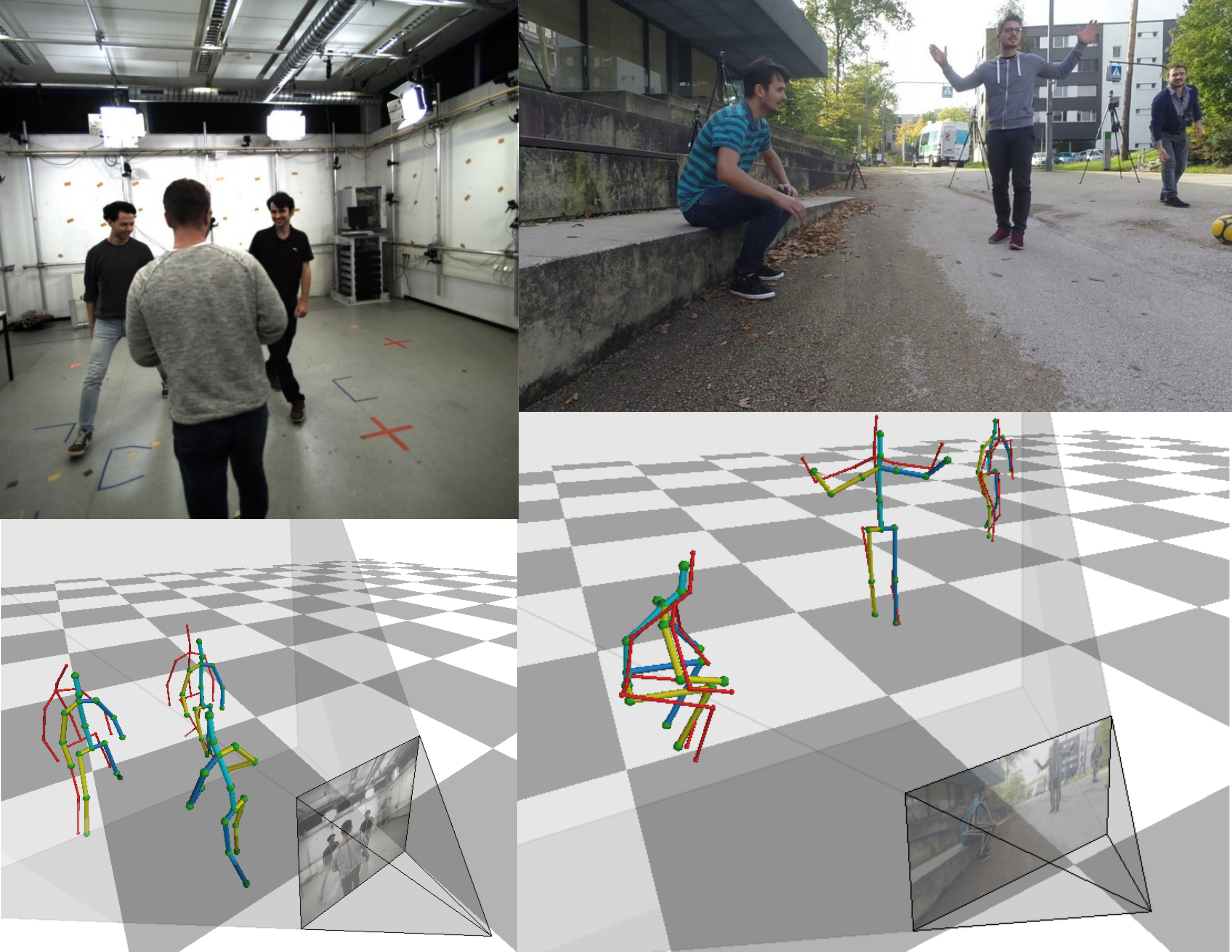} \\
\vspace{-3mm}
\caption{Qualitative results on MuPoTS-3D (prediction in blue-yellow, ground truth in red).}
\label{fig:mupots-qualitative}
\end{figure}

On the multi-person \textbf{MuPoTS-3D}, our MeTRAbs approach yields state-of-the-art results.
For height-normalized skeletons with bone rescaling (standard setting in prior work, Tab.~\ref{tab:mupots}), MeTRAbs outperforms the 2.5D baseline, and the baseline already reaches state-of-the-art results.
Our method performs particularly well on test sequence 2, with heavy occlusions (\eg Fig.~\ref{fig:mupots-qualitative}, left).
Removing the supervision with the absolute 3D loss worsens the absolute PCK of all poses from 38.4\% to 35.0\%.
Surprisingly, the root-relative accuracy seems to improve when turning off the absolute loss.
This is, however, hard to interpret, as Tab.~\ref{tab:mupots} shows an artificial evaluation setting with normalized-height skeletons and bone-rescaling, thereby removing some of the scale recovery aspect from the evaluation.
When evaluating on unnormalized skeletons without bone rescaling (Tab.~\ref{tab:mupots_abs}), it becomes clear that the absolute loss helps: absolute MPJPE improves from 328.8~mm to 248.2~mm, absolute PCK from 36.7\% to 40.2\%, with the root-relative metrics slightly improving as well.
These are state-of-the-art results and improve over methods that are pre- or jointly trained on ground truth pixel-wise depth prediction datasets~\cite{Veges19IJCNN,Veges20Arxiv}.
Further, we can see that the absolute PCK score has high variance and therefore small differences are not necessarily meaningful.
The standard deviation across 5 repeat experiments is around 1.4--3.2\%, and the absolute results for individual test sequences varies strongly as well across different configurations.
This is because the test examples are strongly correlated since they come from video sequences.
Lastly, we note that the detection rate is essentially the same for all of our configurations (Tab.~\ref{tab:mupots_abs}), since we use the same detections, and the official evaluation script performs matching based on the 2D projection, which is very similar across these methods.

In Table~\ref{tab:mupots_persp} we evaluate whether using the full perspective pinhole camera model in the absolute pose reconstruction module brings benefits.
In the last two rows, the absolute loss is not used at training time.
In the other cases we back-propagate the absolute loss gradients either through the weak or full perspective reconstruction method.
We find that training on MuCo with the full perspective model improves the absolute results, but when testing on MuPoTS, it is better to use the weak model.
This may be explained by the fact that people in the MuCo dataset are closer to the camera than in MuPoTS, resulting in stronger perspective effects in MuCo.
To verify this, we computed the ratio of the farthest and closest joint's depth $\max_j{Z_j}/\min_j{Z_j}$ per pose.
If this ratio is large, the weak perspective assumption is a bad approximation.
The median and the 90th percentile of this ratio on MuCo is 1.22 and 1.41, while on MuPoTS it is only 1.16 and 1.26, respectively.
This confirms that perspective effects are stronger in MuCo.

Another possible reason is that the model may output slightly perspective distorted results in the metric 3D head, which are better handled by the weak-perspective model in the next step, as opposed to training time, when the network learns to output the correct metric, perspective-undistorted pose, for which the full perspective model works better afterwards.
Nevertheless, as there is no clear overall winner between the weak and full perspective models, and changing the method across training and test is clearly not desirable, we use the more commonly applied weak perspective method for all other experiments.
\subsection{Inference Speed} Our method is capable of real-time inference.
The root-relative architecture can process 511 crops per second on an RTX 2080~Ti desktop GPU when operating on batches of 8 crops at stride 32 (Tab.~\ref{tab:strides}).
Varying the heatmap resolution using dense prediction provides diminishing returns in accuracy (Tab.~\ref{tab:strides}), showing that soft-argmax can cope with heatmaps of very coarse resolution.
By gathering all person instances of a frame in a batch, MeTRAbs can process 128, 118, 98, 67, 41 frames per second for 1, 2, 4, 8 and 16 people per frame, respectively.
The above calculations assume the image crops are available instantly and the time cost of detection is excluded.
\subsection{ECCV 3DPW Challenge}
Finally, we note that our MeTRAbs method has won the 3D Poses in the Wild (3DPW)~\cite{VonMarcard18ECCV} challenge, organized as a workshop event at the European Conference on Computer Vision, 2020.
Tab. \ref{tab:tdpw} compares results using the 3DPW protocol.
For this, we train our network on a combination of the H3.6M, MuCo, SURREAL~\cite{Varol17CVPR}, SAIL-VOS~\cite{Hu19CVPR} and CMU-Panoptic~\cite{Joo17PAMI} datasets.
We use ResNet-101 as the backbone and additionally apply upper body crop (truncation) augmentation at training time and 5-crop averaging at test time.
When identity tracking is needed, we perform frame-to-frame matching based on absolute pose distance.
The listed methods are not directly comparable due to different training data.
Even with this caveat, our top results show that our approach can scale with further training data and performs well even in challenging in-the-wild scenarios.
\section{Conclusion}
We proposed metric-scale truncation-robust (MeTRo) volumetric heatmaps for the tasks of root-relative and absolute 3D human pose estimation.
These heatmaps directly represent the metric space around the person instead of being tied to the image space and can be predicted with any standard fully-convolutional network.
With a modified weak supervision scheme for 2D labels, careful stride alignment considerations and strong data augmentation, we achieved state-of-the-art results on two important single-person benchmarks: Human3.6M and MPI-INF-3DHP.
We showed that our approach can implicitly discover scale cues from the data, given its superior performance compared to a previous, fixed bone length based heuristic on most test scenarios.
Future research should consider possibilities for learning similar scale cues from large-scale outdoor data as well.
Another interesting future direction can be the evaluation on people with widely differing heights, if such data becomes available on a large scale.
Further, we demonstrated the second benefit of the MeTRo representation, the prediction (``hallucination'') of complete skeletons even when only a part of the body is contained in the image.
For the multi-person absolute 3D pose estimation scenario, we developed a combination of MeTRo heatmaps with 2D heatmap prediction, referred to as MeTRAbs.
We saw the importance of supervising the absolute pose prediction end-to-end by employing a differentiable combination of 2D and root-relative 3D poses.
For this we tested two alternatives, based on weak and full perspective geometry, but neither performed clearly better than the other in our experiments.
Applying MeTRAbs in the top-down multi-person paradigm, we achieved state-of-the-art results on the challenging MuPoTS-3D dataset while keeping the method real-time capable.
Overall we can conclude that heatmap estimation is a versatile paradigm and it is possible to tackle absolute 3D human pose estimation through exclusively estimating heatmaps and encoding all quantities such as coordinates or sizes as activation locations, instead of as activation values.

\section*{Acknowledgments}
We thank Yinglun Liu for help in evaluating on MuPoTS-3D.
This work was funded, in parts, by a Bosch Research Foundation grant, the ERC Consolidator Grant project ``DeeViSe'' (ERC-CoG-2017-773161) and the EU H2020 projects ILIAD (H2020-ICT-2016b-732737) and CROWDBOT (H2020-ICT-2017-779942).
Compute resources were granted by RWTH Aachen University under project ``rwth0479''.

\ifCLASSOPTIONcaptionsoff
  \newpage
\fi

\bibliographystyle{IEEEtran}
\bibliography{IEEEabrv,abbrev_short,references}

\vspace{-10mm}%
\begin{IEEEbiography}[{\includegraphics[width=1in,height=1.25in,clip,keepaspectratio]{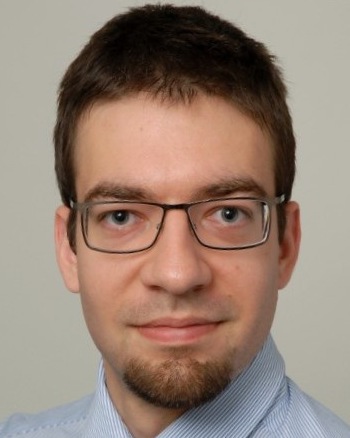}}]{Istv\'an S\'ar\'andi}
is a PhD candidate in the Computer Vision group of RWTH Aachen University, supervised by Prof. Dr. Bastian Leibe. He holds a BSc degree in Computer Engineering from the Budapest University of Technology and Economics, and an MSc degree in Computer Science from RWTH Aachen University.
His research focuses on visual human analysis with an emphasis on 3D body pose for robotics applications. Methods from his first-author publications achieved first place both in the 2018 ECCV PoseTrack Challenge on 3D human pose estimation and in the 2020 ECCV 3D Poses in the Wild Challenge.
\end{IEEEbiography}%
\vspace{-10mm}
\begin{IEEEbiography}[{\includegraphics[width=1in,height=1.25in,clip,keepaspectratio]{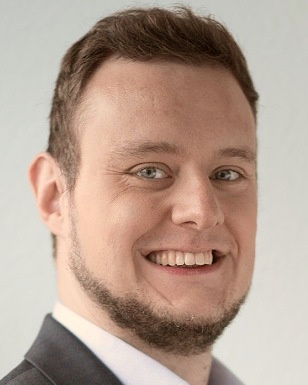}}]{Timm Linder}
defended his PhD thesis on multimodal human detection, tracking and analysis for robots in crowded environments
at the University of Freiburg, Germany in 2020. Since 2016, he is a research scientist in  autonomous systems and robot perception
at Bosch Corporate Research. %
His research interests include computer vision, in particular human detection, tracking and pose estimation, as well as 3D scene generation and sim-to-real transfer.
He has co-authored peer-reviewed publications at major international conferences and journals, served on different program committees in robotics and AI, and received an outstanding reviewer award at ICRA 2019.
\end{IEEEbiography}%
\vspace{-10mm}%
\begin{IEEEbiography}[{\includegraphics[width=1in,height=1.25in,clip,keepaspectratio]{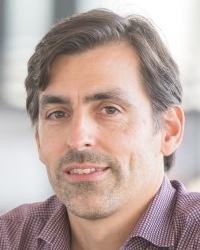}}]{Kai Oliver Arras}
is the head of robotics research and chief expert in robotics at Robert Bosch GmbH. Until 2015, he was assistant professor for social robotics and HRI at the University of Freiburg where he was awarded a DFG Junior Research Group Leader Grant. He obtained his PhD degree from EPFL and was a post-doctoral researcher at KTH Stockholm and at the University of Freiburg. He published around 120 peer-reviewed papers, articles, editorials and book chapters on robot navigation, perception, planning, system integration and was member of various program committees in robotics, AI, HRI and computer vision. %
\end{IEEEbiography}%
\vspace{-10mm}%
\begin{IEEEbiography}[{\includegraphics[width=1in,height=1.25in,clip,keepaspectratio]{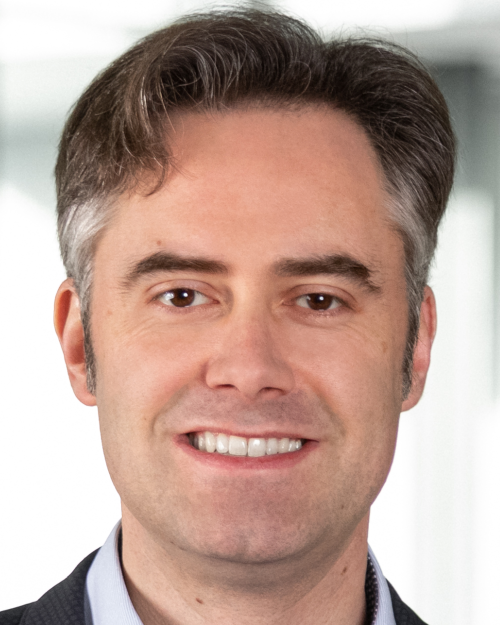}}]{Bastian Leibe}
is a full Professor of Computer Science at RWTH Aachen University, Germany, where he leads the Computer Vision group. He holds an MS degree from the Georgia Institute of Technology (1999), a Diploma degree from the University of Stuttgart (2001), and a PhD from ETH Zurich (2004), all three in Computer Science. His main research interests are in computer vision and machine learning for dynamic visual scene understanding, encompassing object recognition, tracking, segmentation, and 3D reconstruction. He has published over 130 articles in peer-reviewed journals and conferences. Over the years, he has received several awards for his research work, including the CVPR Best Paper Award in 2007, the DAGM Olympus Prize in 2008, and the U.V. Helava Award in 2012. In 2012, he was awarded a European Research Council (ERC) Starting Grant, and in 2017 an ERC Consolidator Grant. He has been Program Chair for ECCV 2016 and Area Chair and program committee member for all major computer vision conferences.
\end{IEEEbiography}

\end{document}